\documentclass[10pt,journal,compsoc]{IEEEtran}
\usepackage{cite}
\usepackage[pdftex]{graphicx}
\usepackage[cmex10]{amsmath}
\usepackage{amssymb}
\usepackage{algorithm,algorithmic}
\usepackage{url,color}

\newcommand{\ie}{\emph{i.e.}}
\newcommand{\eg}{\emph{e.g.}}
\newcommand{\etal}{\emph{et al.}}

\newcommand{\cf}{\emph{cf.}}

\newcommand{\vct}[1]{\ensuremath{\boldsymbol{#1}}} 

\newcommand{\set}[1]{\ensuremath{\mathcal{#1}}}

\begin{document}
\title{Statistical Meta-Analysis of Presentation Attacks for Secure Multibiometric Systems}

\author{Battista~Biggio,~\IEEEmembership{Member,~IEEE,}
        Giorgio~Fumera,~\IEEEmembership{Member,~IEEE,}
        Gian~Luca~Marcialis,~\IEEEmembership{Member,~IEEE,}
        and~Fabio~Roli,~\IEEEmembership{Fellow,~IEEE}
\IEEEcompsocitemizethanks{\IEEEcompsocthanksitem The authors are with the Department of Electrical and Electronic Engineering, University of Cagliari, Piazza d'Armi, 09123 Cagliari, Italy.
\IEEEcompsocthanksitem Battista Biggio: e-mail battista.biggio@diee.unica.it, phone +39 070 675 5776, fax +39 070 675 5782.
\IEEEcompsocthanksitem Giorgio Fumera: e-mail fumera@diee.unica.it, phone +39 070 675 5754, fax +39 070 675 5782.
\IEEEcompsocthanksitem Gian Luca Marcialis: e-mail marcialis@diee.unica.it, phone +39 070 675 5893, fax +39 070 675 5782.
\IEEEcompsocthanksitem Fabio Roli (corresponding author): e-mail roli@diee.unica.it, phone +39 070 675 5779, fax +39 070 675 5782.}}%


\IEEEtitleabstractindextext{%
\begin{abstract}
Prior work has shown that multibiometric systems are vulnerable to presentation attacks,
assuming that their matching score distribution is identical to that of genuine users, without fabricating any fake trait.
We have recently shown that this assumption is not representative of current fingerprint and face presentation attacks, leading one to overestimate the vulnerability of multibiometric systems, and to design less effective fusion rules. 
In this paper, we overcome these limitations by proposing a statistical meta-model of face and fingerprint presentation attacks that characterizes a wider family of fake score distributions, including distributions of known and, potentially, unknown attacks.
This allows us to perform a thorough security evaluation of multibiometric systems against presentation attacks, quantifying how their vulnerability may vary also under attacks that are different from those considered during design, through an uncertainty analysis. 
We empirically show that our approach can reliably predict the performance of multibiometric systems even under never-before-seen face and fingerprint presentation attacks, and that the secure fusion rules designed using our approach can exhibit an improved trade-off between the performance in the absence and in the presence of attack.
We finally argue that our method can be extended to other biometrics besides faces and fingerprints.
\end{abstract}

}

\maketitle

\IEEEdisplaynontitleabstractindextext

%
\IEEEpeerreviewmaketitle

\IEEEraisesectionheading{\section{Introduction}}\label{sect:introduction}

\IEEEPARstart{T}{he} widespread use of biometric identity recognition systems is severely limited by security threats arising from several kinds of potential attacks~\cite{ratha01,jain08,biggio15-spmag}.
In particular, the use of fake biometric traits (\eg, gummy fingers~\cite{matsumoto02}), also known as ``direct'', ``spoofing'' or ``presentation'' attacks~\cite{ratha01,jain08,ISO30107-1,ISO30107-3,ISO2832-37}, 
has the greatest practical relevance as it does not require advanced technical skills. Accordingly, the potential number of attackers is very large.
Vulnerability to presentation attacks has been studied mostly for unimodal fingerprint and face recognition systems~\cite{geller99,matsumoto02,kim09,yang13}.
Multibiometric systems have been considered \emph{intrinsically} more secure, instead, based on the intuition that an attacker should spoof \emph{all} the biometric traits to successfully impersonate the targeted client~\cite{jain04-acm,jain06}.

This claim has been questioned in subsequent work, showing that an attacker may successfully spoof a multibiometric system even if only one (or a subset) of the combined traits is counterfeited, if the fusion rule is not designed by taking that into account~\cite{rodrigues09,rodrigues10,johnson10}.
From a computer security perspective, this is not surprising: it is well-known that the security of a system relies the most on the security of its \emph{weakest link}~\cite{arce03}.
This vulnerability of multibiometric fusion has been shown under the assumption that the fake score distribution at the output of the attacked matcher is identical to that of genuine users, without fabricating any fake trait. Under the same assumption, more secure fusion rules against presentation attacks have also been proposed.

In our recent work, through an extensive experimental analysis, we have shown that the aforementioned assumption is not representative of current face and fingerprint presentation attacks~\cite{biggio11-ijcb,biggio12-iet,fumera14-spoof-chapter}.
In fact, their fake score distributions do not only rarely match those of genuine users, but they can also be very different, depending on the technique, materials, and source images used to fabricate the presentation attack; \ie, presentation attacks can have a different \emph{impact} on the output of the targeted matcher. 
For these reasons, the methodology proposed in~\cite{rodrigues09,rodrigues10,johnson10} may not only provide an overly-pessimistic security evaluation of multibiometric systems to presentation attacks, but also lead one to design secure fusion rules that exhibit a too pessimistic trade-off between the performance in the absence of attack and that under attack.

To perform a more complete analysis of the security of multibiometric systems, as shown in~\cite{biggio11-ijcb,biggio12-iet,fumera14-spoof-chapter,chingovska14-spoof-chapter,hadid15-spmag}, considering only one or a few known attacks is not sufficient. One should thus face the cumbersome and time-consuming task of collecting or fabricating a large, representative set of presentation attacks (with different impact on the attacked matchers) for each biometric characteristic, and evaluate their subsequent effect on the fusion rule and on system security.
However, even in this case, one may not be able to understand how multibiometric systems may behave under scenarios that are different from those considered during design, including presentation attacks fabricated with novel materials or techniques, that may produce \emph{different}, \emph{never-before-seen} fake score distributions at the matchers' output.

In this paper, we propose a methodology for evaluating the security of multibiometric systems to presentation attacks, and for designing secure fusion rules, that overcomes the aforementioned limitations. 
In particular, we start from the underlying idea of~\cite{rodrigues09,rodrigues10,johnson10}, based on  simulating fake score distributions at the matchers' output (without fabricating fake traits), and on evaluating the output of the fusion rule.
However, instead of considering a single fake score distribution equal to that of genuine users, we develop a statistical meta-model that enables simulating a \emph{continuous} family of fake score distributions, with a twofold goal:
($i$) to better characterize the distributions of known, state-of-the-art presentation attacks, as empirically observed in our previous work, through a statistical meta-analysis;
and ($ii$) to simulate distributions that may correspond to never-before-seen (unknown) attacks incurred during system operation, as perturbations of the known distributions, through an uncertainty analysis on the input-output behavior of the fusion rule.
Accordingly, our approach provides both a point estimate of the vulnerability of multibiometric systems against (a set of) known presentation attacks, and an evaluation of how the estimated probability of a successful presentation attack may vary under unknown attacks, giving confidence intervals for this prediction.
To validate the soundness of our approach, we will empirically show that it allows one to model and correctly predict the impact of unknown attacks on system security.
For these reasons, our approach provides a more complete evaluation of multibiometric security. As a consequence, it also allows us to design fusion rules that exhibit a better trade-off between the performance in the absence and in the presence of attack.

The remainder of the paper is structured as follows.
We discuss current approaches for evaluating multibiometric security, and their limitations, in Sect.~\ref{sect:background}.
In Sect.~\ref{sect:meta-analysis}, we present our statistical meta-model of presentation attacks.
In Sect.~\ref{sect:data-model}, we discuss a data model for multibiometric systems that accounts for the presence of zero-effort and spoof impostors (\ie, impostors that do not attempt any presentation attack, and impostors that use at least one fake trait). 
In Sect.~\ref{sect:method} and Sect.~\ref{sect:secure-fusion}, we exploit these models to define a procedure for the security evaluation of multibiometric systems, and to design novel secure fusion rules.
We empirically validate our approach in Sect.~\ref{sect:exp}, using publicly-available datasets of faces and fingerprints that include more recent, never-before-seen presentation attacks, with respect to those considered to develop our meta-model.
Contributions and limitations of our work are finally discussed in Sect.~\ref{sect:conclusions}.

\section{Security of Multibiometric Systems}
\label{sect:background}

Before reviewing current approaches for evaluating multibiometric security against spoofing, it is worth remarking that different terminologies have been used in the literature when referring to biometric systems, and only recently they have been systematized in the development of a novel standard and harmonized vocabulary from the International Standard Organization (ISO)~\cite{ISO30107-1,ISO30107-3,ISO2832-37}. We summarize some of the most common names in Table~\ref{tab:iso-names}, and use them interchangeably in the remainder of this paper.

\begin{figure}[t]
\centering
\includegraphics[width=0.49\textwidth]{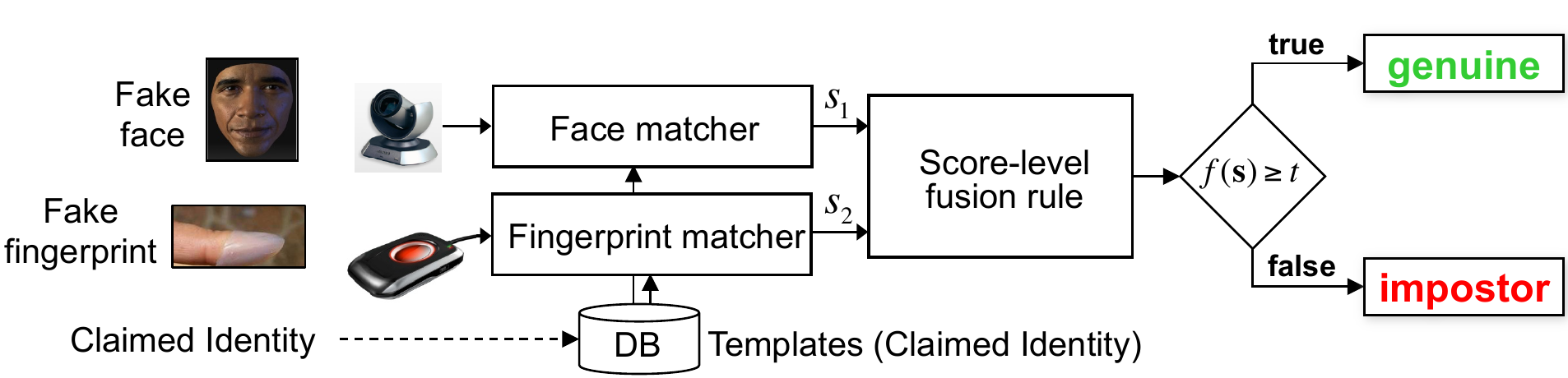}
\vspace{-8pt}
\caption{A bimodal system combining face and fingerprint, that can potentially incur presentation attacks against either biometric, or both.} 
\vspace{-8pt}
\label{fig:multimodal}
\end{figure}

In this work, we focus on multibiometric systems exploiting score-level fusion to combine the matching scores coming from $K$ distinct biometric traits. An example for $K=2$ is shown in Fig.~\ref{fig:multimodal}. During the design phase, authorized clients are enrolled by storing their biometric traits (\ie, templates) and identities in a database. During the online operation, each user provides the requested biometrics, and claims the identity of an authorized client. The corresponding templates are retrieved from the database and matched against the submitted traits. The matching scores $\mathbf s = (s_1, \ldots, s_K) \in \mathbb R^K$ are combined through a fusion rule which outputs an aggregated score $f(\vct s) \in \mathbb R$. The aggregated score is finally compared with a threshold $t$ to decide whether the identity claim is made by a genuine user (if $f(\mathbf s) \geq t$) or an impostor.
Performance is evaluated, as for unimodal systems, by estimating the False Acceptance Rate (${\rm FAR}$) and the False Rejection Rate (${\rm FRR}$) from the genuine and impostor distributions of the aggregated score~\cite{multiBiometrics}.\footnote{Note that, according to the ISO standard~\cite{ISO19795-1}, ${\rm FAR}$ and ${\rm FRR}$ refer to the overall system performance, including errors like failure to capture and failure to extract. If one only considers the algorithmic performance, disregarding these aspects, then the False Match Rate (${\rm FMR}$) and the False Non-Match Rate (${\rm FNMR}$) should be used.}

\begin{table}[tdp]
\caption{ISO standard nomenclature (under development) for biometric systems and presentation attacks~\cite{ISO30107-1,ISO30107-3}, and commonly-used alternatives.}
\vspace{-15pt}
\begin{center}
\begin{tabular}{|p{3.8 cm}|p{4 cm}|} \hline
\textbf{ISO Standard~\cite{ISO30107-1,ISO30107-3}} & \textbf{Commonly-used alternatives} \\ \hline \hline
Artefact(s) & Spoof / Fake Trait(s) \\
Biometric Characteristic(s) & Biometric(s) / Biometric Trait(s) \\
Comparison Subsystem & Matcher / Matching Algorithm\\
Comparison Score(s) & Matching Score(s)\\
Presentation Attack(s) & Spoof / Direct Attack(s) \\
Presentation Attack Detection & Liveness / Spoof Detection \\ \hline
\end{tabular}
\end{center}
\vspace{-20pt}
\label{tab:iso-names}
\end{table}%

Presentation attacks can target any subset of the $K$ biometrics; \eg, a fake face (\eg, a 3D mask) and/or a fake fingerprint can be submitted to the corresponding sensor (see Fig.~\ref{fig:multimodal}). The other impostor's biometrics are submitted to the remaining sensors (if any): such biometrics are said to be subject to a \textit{zero-effort} attack \cite{rodrigues09,biggio12-iet,johnson10,fumera14-spoof-chapter,chingovska14-spoof-chapter}.
In multibiometric systems, the ${\rm FAR}$ is evaluated when all the biometrics are subject to a zero-effort attack, \ie, against zero-effort impostors~\cite{johnson10,chingovska14-spoof-chapter,hadid15-spmag}.
As spoofing attacks affect only the ${\rm FAR}$ of a given system (and not the ${\rm FRR}$), the corresponding performance is evaluated in terms of the so-called Spoof ${\rm FAR}$ (${\rm SFAR}$)~\cite{johnson10,chingovska14-spoof-chapter,hadid15-spmag}. Impostors attempting at least a presentation attack against one of the matchers are referred to as spoof impostors~\cite{johnson10,chingovska14-spoof-chapter}.
Different ${\rm SFAR}$ values can be clearly estimated depending on the combination of attacked matchers, and on the kind of spoofing attacks involved (\eg, one may either use a face mask or a photograph for the purpose of face spoofing).
Furthermore, the ${\rm FAR}$ evaluated against an impostor distribution including both zero-effort and spoof impostors is referred to as Global FAR (${\rm GFAR}$)~\cite{chingovska14-spoof-chapter,hadid15-spmag}.
These measures will be formally defined in Sect.~\ref{sect:data-model}.\footnote{Note that ${\rm SFAR}$ and ${\rm GFAR}$ should not be confused with standard metrics used for presentation attack detection, like the Attack Presentation Classification Error Rate (${\rm APCER}$) and Normal Presentation Classification Error Rate (${\rm NPCER}$)~\cite{ISO30107-1,ISO30107-3,ISO2832-37}, as the latter aim to evaluate the performance of a system that discriminates between alive and fake traits, and not between genuine and impostor users.}

In the following, to keep the notation uncluttered, we will respectively denote
the score distribution of genuine users, zero-effort and spoof impostors at the output of an individual matcher as $p(S^{\rm G})$, $p(S^{\rm I})$ and $p(S^{\rm F})$.

\subsection{Where We Stand Today}
\label{sect:background-soa}

In \cite{rodrigues09}, Rodrigues~\etal{} showed that multibiometric systems can be evaded by spoofing a single biometric characteristic, under the assumption that the corresponding score distribution is identical to that of genuine users:
\begin{equation}
\label{eq:assumption-rodrigues}
p(S^{\rm F}) = p(S^{\rm G}) \, .
\end{equation}
This result was obtained without fabricating any fake trait: the matching scores of presentation attacks were simulated by resampling fictitious scores from $p(S^{\rm G})$. Similar results were obtained under the same assumption in~\cite{rodrigues10,johnson10}.

Subsequently, we carried out an extensive experimental analysis on multibiometric systems combining face and fingerprint, to evaluate whether and to what extent this assumption was representative of current techniques for fabricating fake traits (from now on, \textit{fake fabrication techniques})~
\cite{biggio12-iet,biggio11-ijcb,fumera14-spoof-chapter}.\footnote{For example, fake fingerprints can be fabricated adopting a mold of plasticine-like material where the targeted client has put his own finger, and a cast of gelatin dripped over the mold; fake faces can be fabricated by printing a picture of the targeted client on paper.}
To this end, we considered different acquisition sensors and matchers, and several presentation attacks fabricated with different techniques.
We used five state-of-the-art fake fabrication techniques for fingerprints, as listed in Table~\ref{tab:fakedata}, taken from the first two editions of the Fingerprint Liveness Detection Competition (LivDet)~\cite{marcialis09,livdet11}. They include fake traits fabricated with five different materials,
in a worst-case setting commonly accepted for security evaluation of fingerprint recognition systems, \ie, with the cooperation of the targeted client (\emph{consensual} method).
Fake faces (see Table~\ref{tab:fakedata}) were obtained by displaying a photo of the targeted client on a laptop screen, or by printing it on paper, and then showing it to the camera \cite{chakka11,zhang11}. Pictures were taken with cooperation (\emph{Print} \cite{anjos11,chakka11} and \emph{Photo Attack} datasets \cite{biggio11-ijcb}) and without cooperation of the targeted client (\emph{Personal Photo Attack} \cite{biggio11-ijcb}). As a fourth fake fabrication technique, wearable masks were also fabricated to impersonate the targeted clients (\emph{3D Mask Attack} \cite{erdogmus13-btas}).

Our analysis showed that most of the above fake fabrication techniques produced distributions $p(S^{\rm F})$ that did not match the overly-pessimistic assumption of \cite{rodrigues09}.
Representative examples are reported in Fig.~\ref{fig:spoof-distributions} for several unimodal fingerprint and face systems, where it can be seen that: $p(S^{\rm F})$ is located in between $p(S^{\rm G})$ and $p(S^{\rm I})$, it can be very different from $p(S^{\rm G})$, and its shape depends on the fake fabrication technique (see the three rightmost plots in the top row of Fig.~\ref{fig:spoof-distributions}, which refer to presentation attacks against the same fingerprint system with three different fake fabrication techniques). 
Notably, this is also true for more recent work on presentation attacks, even against different biometrics, like palm vein and iris~\cite{tome15-icb,busch15-tifs}.
Accordingly, the ${\rm SFAR}$ estimated under the assumption in~\cite{rodrigues09} can be much higher than the actual ${\rm SFAR}$~\cite{biggio12-iet,biggio11-ijcb,fumera14-spoof-chapter}.

Under the viewpoint of security evaluation, the approach proposed in~\cite{rodrigues09,rodrigues10,johnson10} has the advantage of avoiding the fabrication of fake traits. However, it can give one an overly-pessimistic and incomplete analysis of multibiometric security, as $p(S^{\rm F})$ only rarely matches $p(S^{\rm G})$, and different presentation attacks produce different $p(S^{\rm F})$.

\begin{table}[t] 
\caption{Dataset characteristics: number of clients (\# clients), and number of spoof (\# spoofs) and live (\# live) images per client.}
\vspace{-20pt}
\begin{center}  
\begin{tabular}{| l | l| l | l | }
\hline
Dataset & \# clients & \# spoofs & \# live \\ 
\hline \hline
LivDet09-Silicone (catalyst)~\cite{marcialis09} & 142 & 20 & 20 \\
LivDet11-Alginate~\cite{livdet11} & 80 & 3 & 5 \\
LivDet11-Gelatin~\cite{livdet11} & 80 & 3 & 5 \\
LivDet11-Silicone~\cite{livdet11} & 80 & 3 & 5 \\
LivDet11-Latex~\cite{livdet11} & 80 & 3 & 5 \\
\hline
Photo Attack~\cite{biggio11-ijcb}  & 40 & 60 & 60\\
Personal Photo Attack~\cite{biggio11-ijcb} & 25 & 5 (avg.)& 60\\
Print Attack~\cite{anjos11,chakka11}  & 50 & 12 & 16 \\
3D Mask Attack~\cite{erdogmus13-btas} & 17 & 5 (video) & 10 (video) \\
\hline
\end{tabular}
\label{tab:fakedata}
\vspace{-10pt}
\end{center}
\end{table}

\subsection{Limitations and Open Issues}
\label{sect:background-limitations}

The empirical evidences summarized in Sect.~\ref{sect:background-soa} highlight that, even if the assumption in~\cite{rodrigues09,rodrigues10,johnson10} allows us to assess the ${\rm SFAR}$ of a multibiometric system under different combinations of attacked matchers \emph{without fabricating any fake trait}, the corresponding estimates may be too pessimistic. For the same reason, secure fusion rules based on the same assumption may even worsen the trade-off between the performance in the absence and in the presence of spoofing.

In addition, these evaluation techniques, as well as more recent ones~\cite{chingovska14-spoof-chapter,hadid15-spmag}, do not specifically account for never-before-seen presentation attacks that may be incurred during system operation.
System security is often evaluated on a set of attacks that may not be representative enough of future ones, \ie, considering fake score distributions that may be very different from those exhibited by unknown attacks. Accordingly, such evaluations only provide point estimates for the corresponding ${\rm SFAR}$, without giving any information on how it may vary under presentation attacks that are different from those considered during design.

This raises two main open issues:
($i$) how to perform a more complete security evaluation of multibiometric systems, accounting for both known and unknown presentation attacks;
and ($ii$) how to design improved secure fusion rules,
while avoiding the cumbersome task of collecting or fabricating a large set of representative fake biometric traits.

In the next section, we present a statistical meta-model of presentation attacks that overcomes the aforementioned limitations by simulating a \emph{continuous} family of fake score distributions $p(S^{\rm F})$, including distributions that correspond to known presentation attacks, and potential variations to account for never-before-seen attacks.
We will then discuss how to exploit our meta-model for the purpose of security evaluation and to design secure fusion rules (Sects.~\ref{sect:method}-\ref{sect:secure-fusion}).

\section{Statistical Meta-Analysis of Presentation Attacks}
\label{sect:meta-analysis}

To address the issues discussed in Sect.~\ref{sect:background-limitations}, in this section we develop a statistical meta-model that allows one to simulate a \emph{continuous} family of fake score distributions (at the output of a matcher), with a twofold goal: ($i$) to characterize the score distributions of known, state-of-the-art presentation attacks; and ($ii$) to simulate continuous perturbations of such distributions that may correspond to the effect of unknown presentation attacks.
In both cases, no fabrication of presentation attacks is required, since the corresponding fake score distributions are simulated.

To this aim, we exploit two well-known techniques of statistical data analysis:: ($i$) statistical meta-analysis, that aims to find common patterns from results collected from previous studies, to avoid repeating time-consuming and costly experiments (see, \eg,~\cite{sohn99-tpami}); and ($ii$) uncertainty analysis, to evaluate the output of a system under unexpected input perturbations (see, \eg, \cite{saltelli00,saltelli00-book,kann00}).
In particular, the latter allows us to investigate the input-output behavior of fusion rules, accounting for perturbations of the fake score distributions at the matchers' output (\ie, inputs to the fusion rule) that may be caused by unknown attacks. This completes our evaluation by providing information on how the system may behave under presentation attacks that are different from those considered during design.

Accordingly, our approach may provide a point estimate of the vulnerability of multibiometric systems against known presentation attacks (\eg, a given ${\rm SFAR}$ value), as current evaluation techniques~\cite{rodrigues09,rodrigues10,johnson10,chingovska14-spoof-chapter,hadid15-spmag},
and also an estimate of how this vulnerability may vary under unknown attacks, giving confidence intervals for this prediction. As we will show experimentally, this provides a more thorough security evaluation of multibiometric systems, capable of correctly predicting even the impact of never-before-seen presentation attacks.

In the following we present our meta-model, and show how it characterizes known presentation attacks.


\subsection{A Meta-model of Presentation Attacks}
\label{sect:alpha-model}

\begin{figure*}[tb]
\begin{center}
\includegraphics[width = 0.24\textwidth]{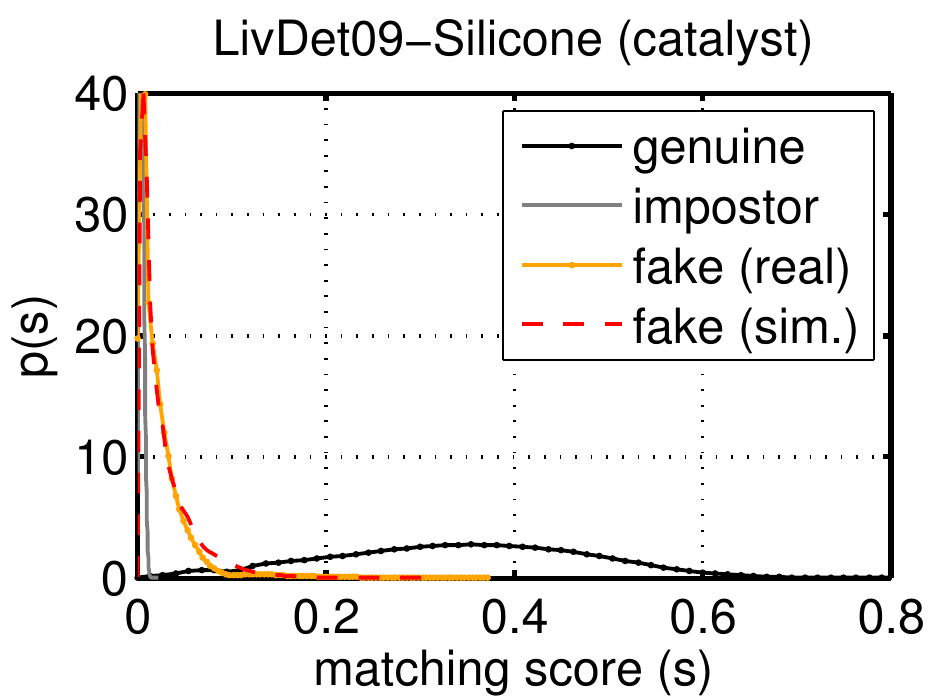}
\includegraphics[width = 0.24\textwidth]{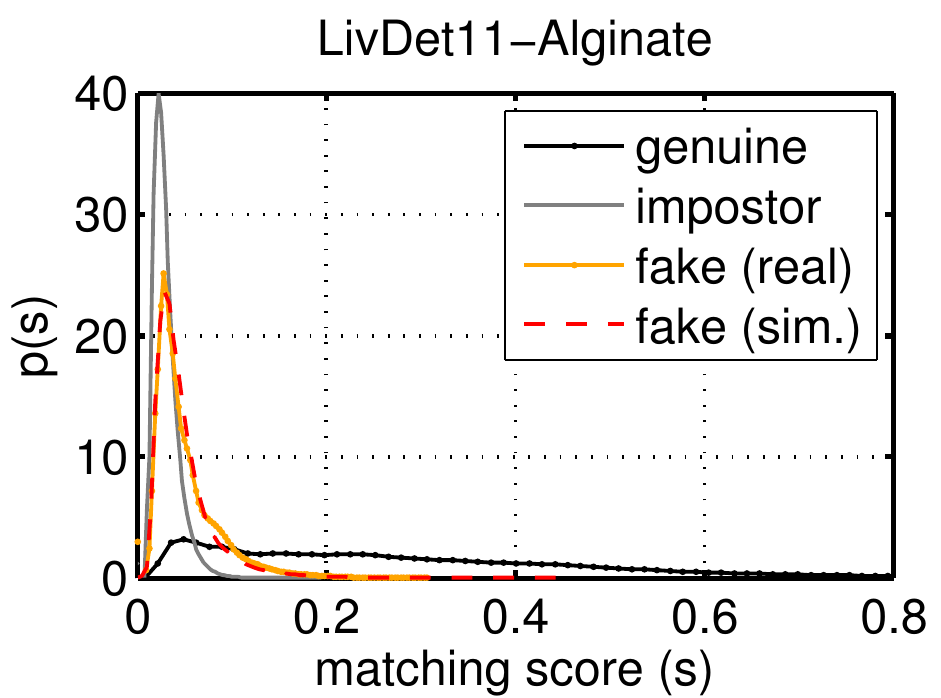}
\includegraphics[width = 0.24\textwidth]{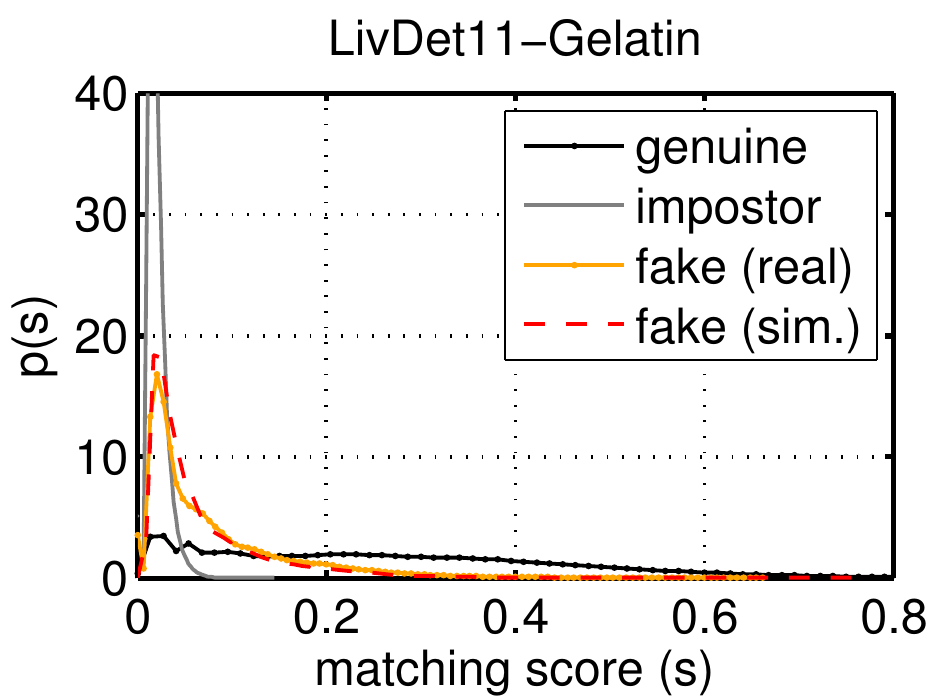}
\includegraphics[width = 0.24\textwidth]{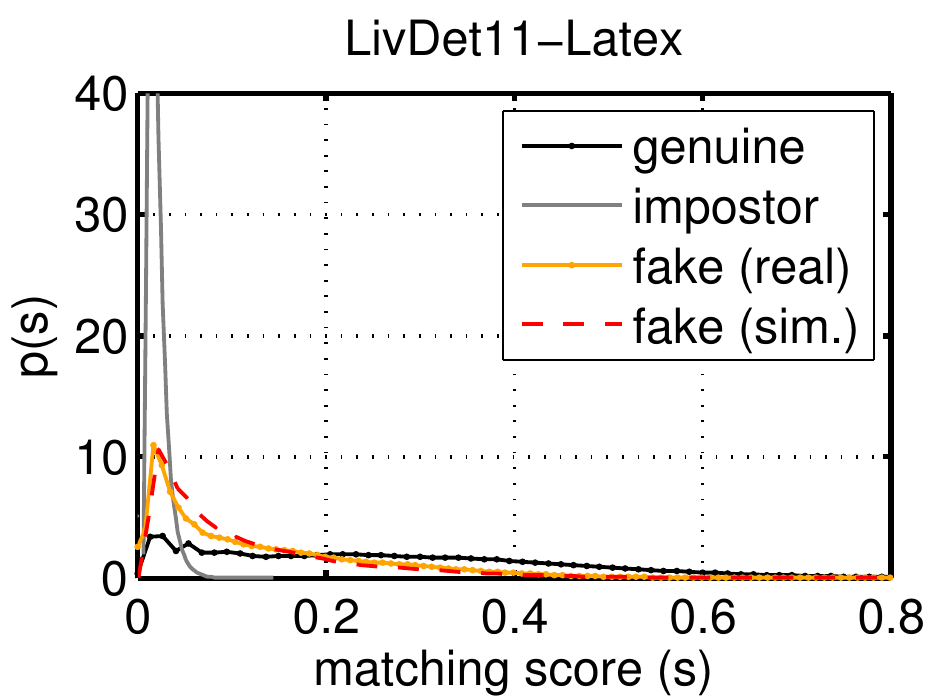}\\
\includegraphics[width = 0.24\textwidth]{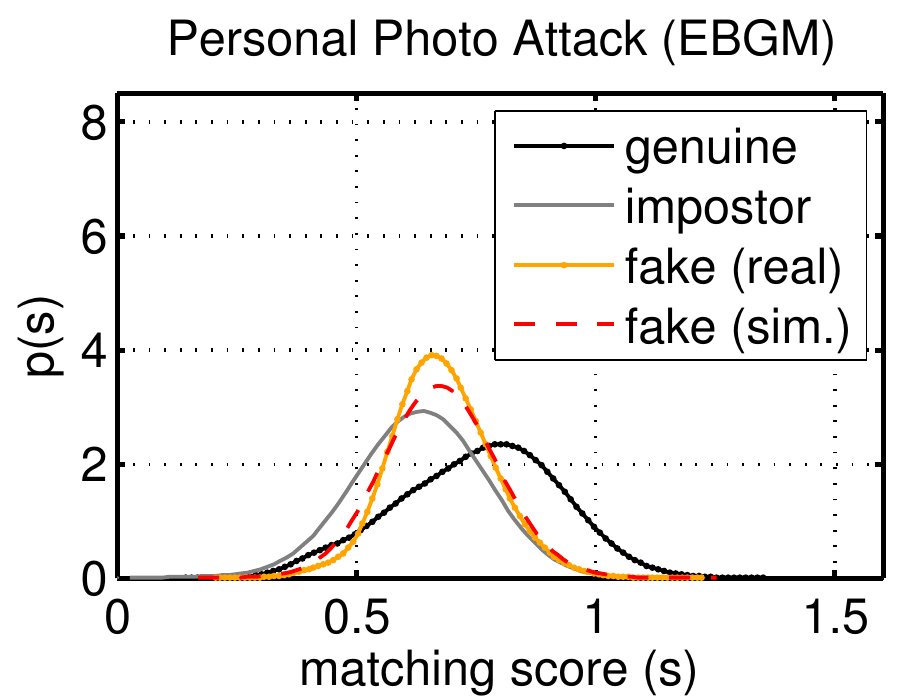}
\includegraphics[width = 0.24\textwidth]{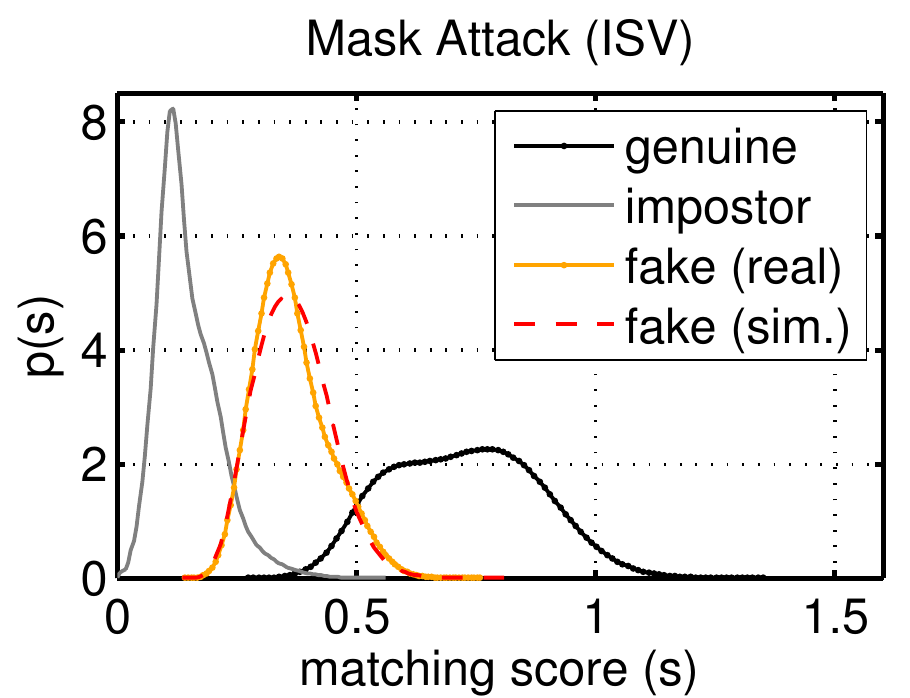}
\includegraphics[width = 0.24\textwidth]{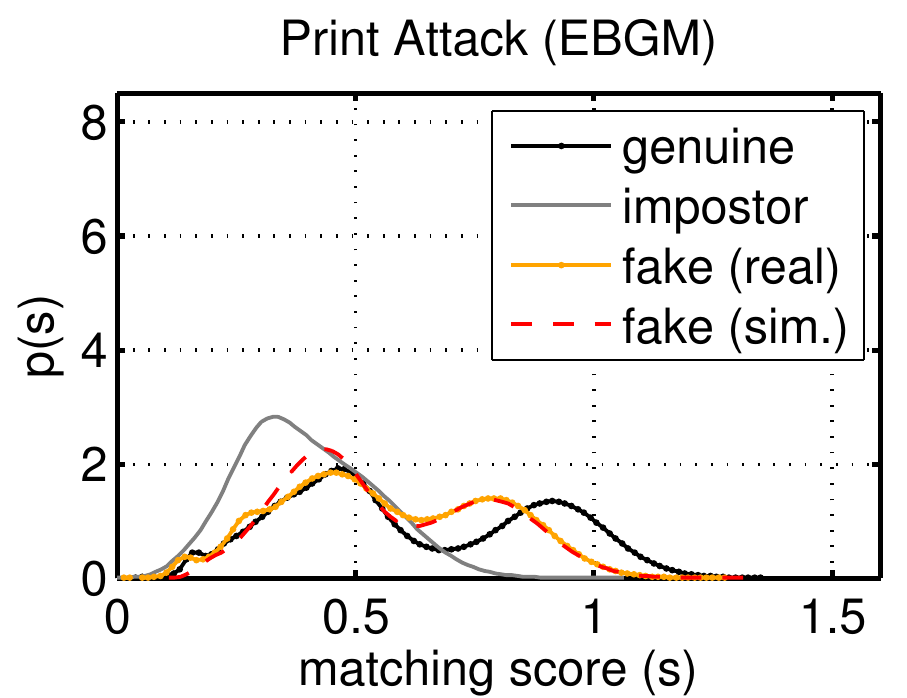}
\includegraphics[width = 0.24\textwidth]{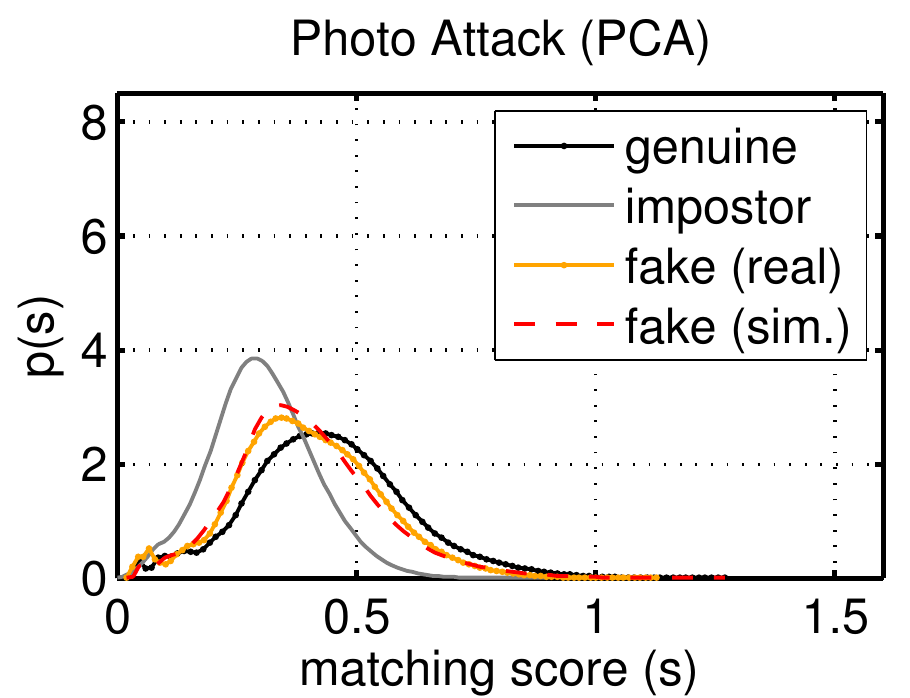}
\vspace{-8 pt}
\caption{Matching score distributions of fake fingerprints (top row) and faces (bottom row)
for
LivDet09-Silicone (\emph{Sensor}: Biometrika, \emph{Matcher}: Veryfinger),
LivDet11-Alginate (\emph{S}: Biometrika, \emph{M}: Bozorth3),
LivDet11-Gelatin (\emph{S}: Italdata, \emph{M}: Bozorth3),
LivDet11-Latex (\emph{S}: Italdata, \emph{M}: Bozorth3),
Personal Photo Attack (\emph{S}: commercial webcam, \emph{M}: EBGM),
3D Mask Attack (\emph{S}: Microsoft Kinect, \emph{M}: ISV),
Print Attack (\emph{S}: commercial webcam, \emph{M}: EBGM),
Photo Attack (\emph{S}: commercial webcam, \emph{M}: PCA).
Simulated fake distributions according to our model are also shown for comparison, at \emph{low risk} (LivDet09-Silicone, LivDet11-Alginate, 3D Mask Attack, and  Personal Photo Attack), \emph{medium risk} (LivDet11-Gelatin, and Print Attack), and \emph{high risk} (LivDet11-Latex, and Photo Attack). The values of $(\mu_{\alpha},\sigma_{\alpha})$ used to simulate each attack scenario are reported in Table~\ref{tab:model-parameters}.} 
\label{fig:spoof-distributions}
\vspace{-15 pt}
\end{center}
\end{figure*}
We first carry out a statistical meta-analysis of the fake score distributions of known presentation attacks described in Sect.~\ref{sect:background-soa}.
We already observed that they exhibit very different shapes across the different multibiometric systems and fake fabrication techniques considered. Nevertheless, under a closer scrutiny some general patterns emerge (see Fig.~\ref{fig:spoof-distributions}).
First, $p(S^{\rm F})$ always lies in between $p(S^{\rm I})$ and $p(S^{\rm G})$~\cite{biggio12-iet,biggio11-ijcb,erdogmus13-btas,fumera14-spoof-chapter}.
Second, it exhibits a clear similarity either to $p(S^{\rm I})$ or to $p(S^{\rm G})$, or an intermediate shape between them.
This is reasonable, since presentation attacks attempt to mimic the clients' traits: accurate reproductions are likely to result in a score distribution more similar to $p(S^{\rm G})$, whereas inaccurate ones (provided that they still ``resemble'' a real trait to the sensor/matcher) are likely to yield a score distribution more similar to $p(S^{\rm I})$.
In particular we observed that, for fingerprint spoofing, $p(S^{\rm F})$ resembles $p(S^{\rm I})$ for lower scores, and $p(S^{\rm G})$ for higher scores, exhibiting a significant \emph{heavy tail} behavior (see Fig.~\ref{fig:spoof-distributions}, top row); in the case of face spoofing, it exhibits a shape intermediate between $p(S^{\rm I})$ and $p(S^{\rm G})$, without significant heavy tails (see Fig.~\ref{fig:spoof-distributions}, bottom row).

To reproduce the above patterns, we propose a meta-model of $p(S^{\rm F})$ as a mixing of $p(S^{\rm I})$ and $p(S^{\rm G})$, by defining a \emph{fictitious} r.v.~$S^{\rm F}$ as:
\begin{equation}
S^{\rm F} = \alpha S^{\rm G} + (1 - \alpha) S^{\rm I} \ ,
\label{eq:alpha-model}
\end{equation}
where we define in turn $S^{\rm G}$, $S^{\rm I}$ and $\alpha$ as independent r.v.~drawn respectively from the empirical distributions $p(S^{\rm G})$ and $p(S^{\rm I})$,\footnote{Note that, assuming independence between $S^{\rm G}$ and $S^{\rm I}$ here is only used to generate the fictitious scores $S^{\rm F}$, and neither requires nor assumes independence between the genuine and impostor distributions.}
and a distribution
${\rm Beta}(\alpha ; \mu_{\alpha}, \sigma_{\alpha})$,
with mean $\mu_{\alpha} \in (0,1)$ and standard deviation $\sigma_{\alpha} \in (0, \sqrt{\mu_{\alpha}(1-\mu_{\alpha})})$.\footnote{For the sake of interpretability, we consider the mean and standard deviation of a Beta distribution, instead of the parameters $p,q>0$.}
Note that the meta-model \eqref{eq:alpha-model} is more flexible than a standard mixture, as $\alpha$ itself is a r.v.

The choice of a Beta distribution is motivated by the fact that a Beta r.v.~ranges in $[0,1]$, which allows the resulting $p(S^{\rm F})$ to exhibit the pattern mentioned above, \ie, an intermediate shapes between $p(S^{\rm G})$ and $p(S^{\rm I})$.
As limit cases, when $p(\alpha=0) = 1$, one obtains $p(S^{\rm F})=p(S^{\rm I})$, and when $p(\alpha=1) = 1$ one obtains $p(S^{\rm F})=p(S^{\rm G})$.
In all the other cases, the achievable ${\rm SFAR}$ values, for any fixed decision threshold, range between those of $p(S^{\rm F})=p(S^{\rm I})$ and $p(S^{\rm F})=p(S^{\rm G})$.
The above limit distributions correspond therefore to the worst- and best-case presentation attack for a biometric system, in terms of the corresponding ${\rm SFAR}$, that our meta-model can represent.
These are reasonable choices, since $p(S^{\rm F})=p(S^{\rm G})$ can be seen as a worst-case attack, and correspond to the attack scenario of\cite{rodrigues09,rodrigues10,johnson10} (see Eq.~\ref{eq:assumption-rodrigues});
and $p(S^{\rm F})=p(S^{\rm I})$ corresponds to the least harmful spoofing attack of interest for security evaluation, since attacks leading to a lower ${\rm SFAR}$ are by definition less effective than a zero-effort attack, which is already included in the standard evaluation of biometric systems.

Note that, if we set $\sigma_{\alpha} = 0$, the meta-model \eqref{eq:alpha-model} becomes equivalent to the one we proposed in \cite{akhtar12-icb}, where $\alpha$ was defined as a \emph{constant} parameter (not a r.v.) ranging in $[0,1]$. Although this is a simpler model, it turned out to be not flexible enough to properly fit all the fake score distributions empirically observed in our subsequent work \cite{biggio12-iet,biggio11-ijcb,fumera14-spoof-chapter}.

\textbf{Inferring the meta-parameters}.
Given a set of genuine and impostor scores obtained, for a given trait, in a laboratory setting, and a set of scores obtained from fake traits fabricated with a given technique, fitting the empirical distribution $p(S^{\rm F})$ with our meta-model~\eqref{eq:alpha-model} amounts to inferring the value of the meta-parameters $(\mu_{\alpha}, \sigma_{\alpha})$.
To this aim the \emph{method of moments} can be exploited.\footnote{An alternative is a log-likelihood maximization approach, which is however more computationally demanding.
}
In our case it simply amounts to setting $\mu_{\alpha}$ and $\sigma_{\alpha}$ to the values that can be analytically obtained from Eq.~\eqref{eq:alpha-model}, as a function of means and variances of the distributions of genuine, impostor and fake scores, estimated from the available data:
\begin{eqnarray}
\label{eq:fit-mean-alpha}
\mu_{\alpha} & = & \frac{\mu_{S^{\rm F}} - \mu_{S^{\rm I}}}{  \mu_{S^{\rm G}}-\mu_{S^{\rm I}} } \, ,\\
\label{eq:fit-std-alpha}
\sigma_{\alpha} &=& \sqrt{\frac{\sigma^{2}_{S^{\rm F}} -  \mu^{2}_{\alpha} \sigma^{2}_{S^{\rm G}} -
(1-\mu_{\alpha})^{2} \sigma^{2}_{S^{\rm I}}}{\sigma^{2}_{S^{\rm G}} +
 \sigma^{2}_{S^{\rm I}} +
(\mu_{S^{\rm G}} - \mu_{S^{\rm I}})^{2} }}\,.
\end{eqnarray}
Note however that the estimated values of $\mu_{\alpha}$ and  $\sigma_{\alpha}$ may not satisfy the constraints $\mu_{\alpha} \in (0,1)$ and $\sigma_{\alpha} \in (0,\sqrt{\mu_{\alpha}(1-\mu_{\alpha})})$ that must hold for a Beta distribution. In this case, one should correct the corresponding estimate(s) to the closest (minimum or maximum) admissible value.

To sum up, our meta-model characterizes a known fake score distribution with given values of the meta-parameters $\mu_{\alpha}$ and $\sigma_{\alpha}$.
The associated presentation attack can then be simulated on \emph{any} multibiometric system that uses the same biometric trait, by simulating the fake score distribution using Eq.~\eqref{eq:alpha-model}, with the given values of $\mu_{\alpha}$ and $\sigma_{\alpha}$, and the corresponding empirical distributions $p(S^{\rm G})$ and $p(S^{\rm I})$.

Finally, our meta-model \eqref{eq:alpha-model} can be exploited also to carry out an uncertainty analysis.
This can be attained by considering simulated fake score distributions obtained by perturbing the values of the $\mu_{\alpha}$ and $\sigma_{\alpha}$ parameters associated to the known presentation attacks.
Note that this produces a \emph{continuous} family of simulated distributions.

The exact security evaluation procedure we propose is described in Sect.~\ref{sect:method}.
In the following we analyze how our meta-model characterizes known presentation attacks.

\subsection{Attack Scenarios for Fingerprints and Faces}
\label{sect:attack-scenario-definition}

\begin{figure}[tb]
\begin{center}
\includegraphics[width = 0.43\textwidth]{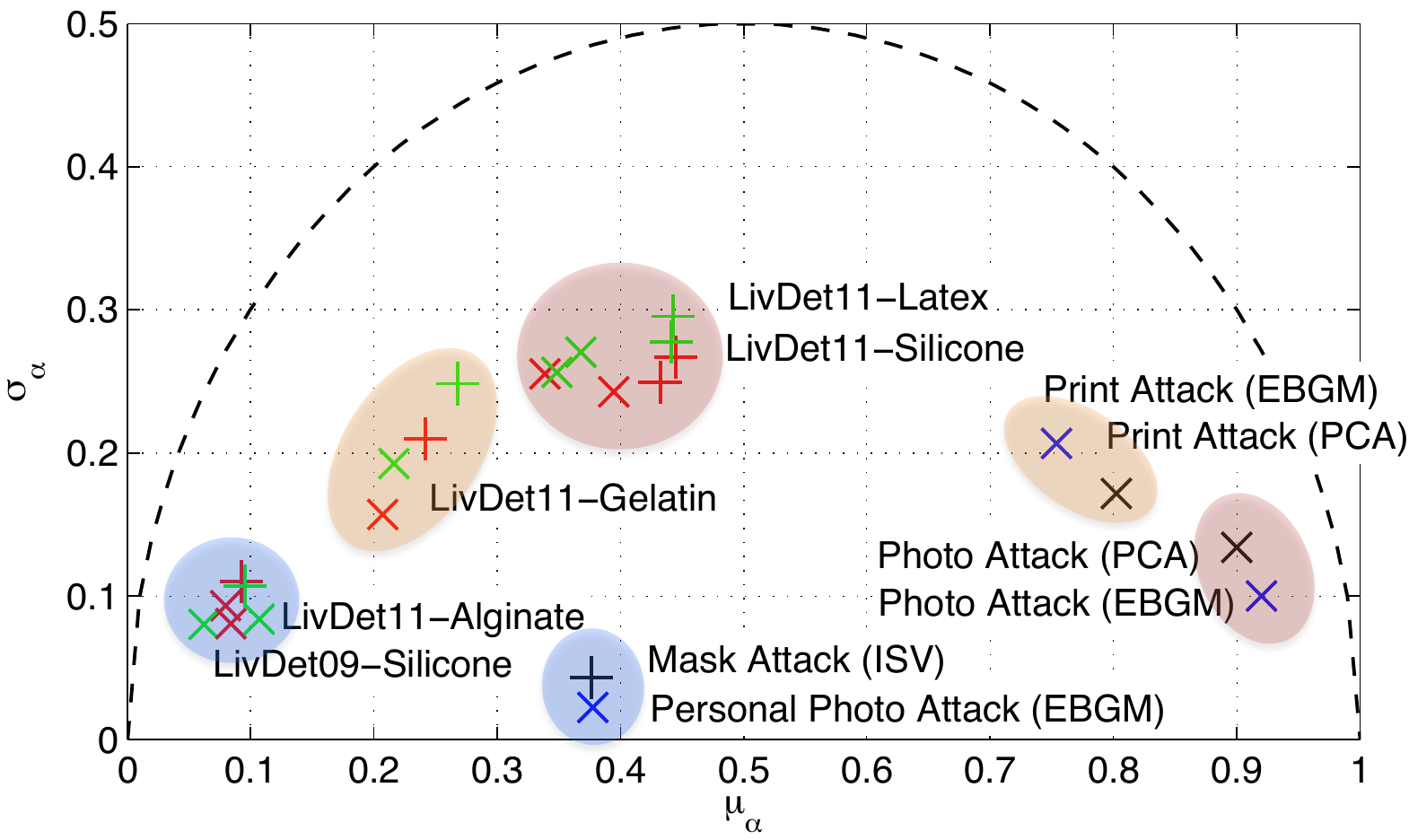}
\vspace{-8pt}
\caption{Results of fitting our model to the considered datasets \cite{biggio12-iet,biggio11-ijcb,erdogmus13-btas}.
Each real fake distribution is represented as a point with coordinates $(\mu_{\alpha}, \sigma_{\alpha})$.
A red (green) `x' (`+') denotes a fake fingerprint distribution obtained by the Bozorth3 (Veryfinger) matcher and the Biometrika (Italdata) sensor. A blue (black) `x' denotes a fake face distribution obtained by the EBGM (PCA) matcher and a commercial webcam. The black `+' denotes the distribution of fake faces for the 3D Mask Attack database, obtained by the ISV matcher and the Microsoft Kinect sensor.
The area under the dashed black curve corresponds to $\sigma_{\alpha} \leq \sqrt{\mu_{\alpha}(1-\mu_{\alpha})}$, and delimits the family of possible fake distributions generated by our meta-model.
Low-, medium-, and high-impact presentation attacks clustered to form the corresponding attack scenarios are highlighted respectively as blue, orange, and red shaded areas.}
\label{fig:model-fitting}
\end{center}
\vspace{-15pt}
\end{figure}

Fig.~\ref{fig:spoof-distributions} shows some representative examples of the distributions $p(S^{\rm I})$, $p(S^{\rm G})$ and $p(S^{\rm F})$ collected for our statistical meta-analysis, together with the fitting distributions of $p(S^{\rm F})$ obtained with our meta-model as explained above.
In all cases the fitting accuracy turned out to be very good.
The values of $(\mu_{\alpha}, \sigma_{\alpha})$ obtained for all datasets, including those of Fig.~\ref{fig:spoof-distributions}, are shown in Fig.~\ref{fig:model-fitting}, where each point corresponds to a specific Beta distribution of the r.v.~$\alpha$.
Note that the closer the points in the $\mu_{\alpha}, \sigma_{\alpha}$ plane, the closer the corresponding Beta distributions, and thus the distribution $p(S^{\rm F})$, for a given pair $p(S^{\rm G})$, $p(S^{\rm I})$.
The different shapes exhibited by the fake score distributions (see, \eg, Fig.~\ref{fig:spoof-distributions}) are mirrored by the different values of the corresponding meta-parameters, which are spread over a considerable region of the $(\mu_{\alpha}, \sigma_{\alpha})$ plane (see Fig.~\ref{fig:model-fitting}).
Note also that, often, $\sigma_{\alpha} > 0$: this explains why the simpler model we proposed in \cite{akhtar12-icb}, corresponding to $\sigma_{\alpha}=0$, does not provide a good fitting in such cases, as mentioned above. In particular, it can not properly characterize the heavy tails exhibited by the score distributions of fake fingerprints (see Fig.~\ref{fig:spoof-distributions}).

\begin{figure}[tb]
\centering
\includegraphics[width = 0.43\textwidth]{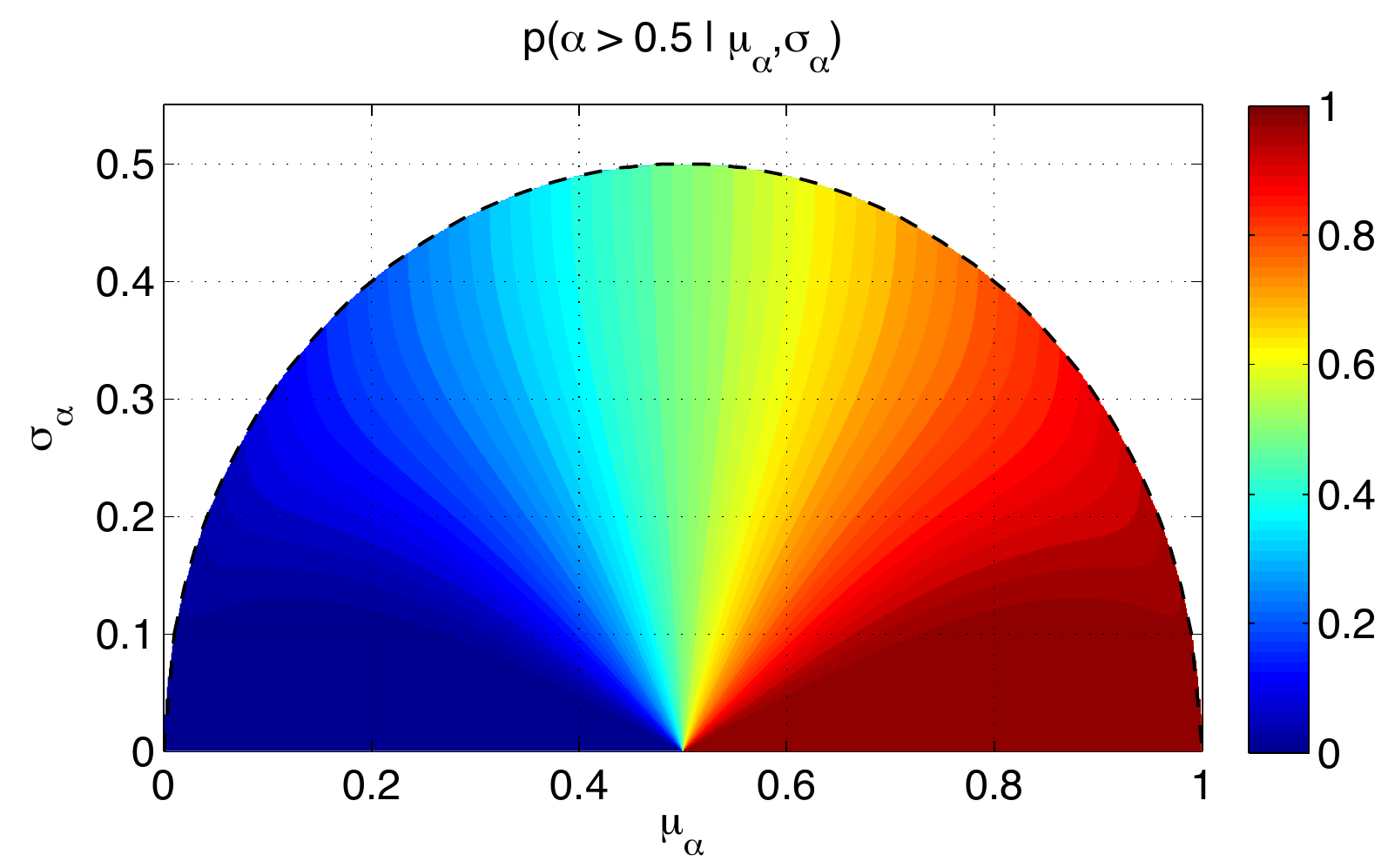}
\vspace{-12pt}
\caption{Attack impact for each attack scenario of our meta-model (Eq.~\ref{eq:risk-level}).}
\label{fig:risk}
\vspace{-16pt}
\end{figure}

In Sect.~\ref{sect:background}, we observed that some attacks produce fake score distributions that are ``closer'' to the score distribution of genuine users, rather than to that of impostors (\eg, our fingerprint replicas made with latex are much more similar to the ``alive'' ones than replicas made with silicone).
A useful by-product of our meta-model is that it allows us to formalize the above qualitative observation, by quantifying the \emph{impact} of each presentation attack in terms of the probability that it is successful.
To this aim, it is not possible to directly use the ${\rm SFAR}$ value computed for some given decision threshold, as the fake score distribution simulated by our meta-model \eqref{eq:alpha-model} depends also on the distributions $p(S^{\rm G})$ and $p(S^{\rm I})$.
Nevertheless, for any given $p(S^{\rm G})$ and $p(S^{\rm I})$, the more the $\alpha$ distribution is concentrated towards higher values, the closer the corresponding $p(S^{\rm F})$ is to $p(S^{\rm G})$, and thus the higher the ${\rm SFAR}$ for any given decision threshold.
Accordingly, we propose to quantify the impact of an attack associated to given values of $(\mu_{\alpha}, \sigma_{\alpha})$ as:
\begin{equation}
\label{eq:risk-level}
p(\alpha > 0.5 | \mu_{\alpha}, \sigma_{\alpha}) \ ,
\end{equation}
where $0.5$ is a reasonable value to evaluate how much the Beta distribution is concentrated towards high values, as $\alpha \in [0,1]$.\footnote{We argue that this measure of the \emph{attack impact} may be also useful to quantitatively evaluate some aspects of the \emph{attack potential}, a metric under definition in~\cite{ISO30107-3}. This can be investigated in future work.}
Fig.~\ref{fig:risk} shows the values of Eq.~\eqref{eq:risk-level} for each point of the $(\mu_{\alpha}, \sigma_{\alpha})$ plane.
Note that known fake score distributions exhibit very different values of the \emph{attack impact}.

\begin{figure}[t]
\begin{center}
\includegraphics[width = 0.41\textwidth]{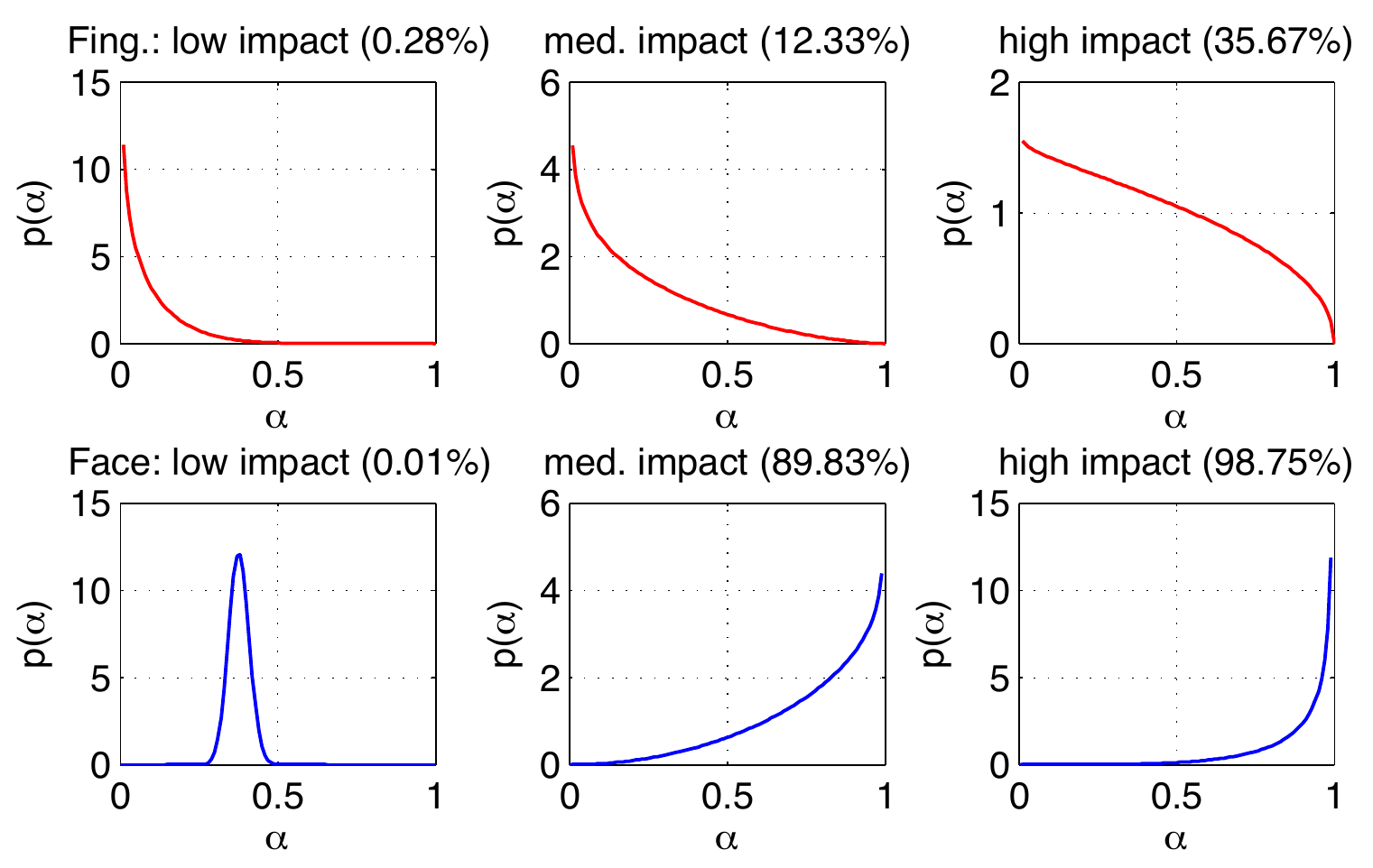}
\vspace{-12pt}
\caption{Beta distributions and attack impact (in parentheses) for the three fingerprint and face presentation attack scenarios identified in Fig.~\ref{fig:model-fitting}.}
\vspace{-13pt}
\label{fig:beta-attack-scenarios}
\end{center}
\end{figure}

Another interesting by-product emerging from our statistical meta-analysis is that score distributions produced by the same fake fabrication technique (\eg, fake fingerprints fabricated with latex) are fitted by our meta-model with very similar $(\mu_{\alpha}, \sigma_{\alpha})$ values, across \emph{different} sensors, matchers, and user populations (\ie, different $p(S^{\rm G})$ and $p(S^{\rm I})$); on the other hand, the same sensor, matcher, and user population can result in considerably different $(\mu_{\alpha}, \sigma_{\alpha})$ values for different fake fabrication techniques (\eg, fake fingerprints fabricated with alginate, gelatin and latex).
This implies that the attack impact measure of Eq.~\ref{eq:risk-level} mainly depends on the kind of attack, and is almost independent on the specific multibiometric system.
Accordingly, our meta-model allows one to quantitatively compare the impact of different kinds of presentation attacks, either against the same or different multibiometric systems.
The above result also means that, for both considered traits, our meta-model produces compact clusters of $(\mu_{\alpha}, \sigma_{\alpha})$, each representing fake score distributions associated to one or more different fake fabrication techniques. These clusters are highlighted in Fig.~\ref{fig:model-fitting}.
This result allows one to use a \emph{single} instantiation of our meta-model for approximating all the distributions $p(S^{\rm F})$ corresponding to the fake fabrication technique(s) lying in the same cluster, encompassing all the underlying, different multibiometric systems. For instance, the corresponding pair of $(\mu_{\alpha}, \sigma_{\alpha})$ values can be defined as the cluster centroid, with no appreciable loss in fitting accuracy.
In particular, we can identify in the considered data the three clusters for fingerprint spoofing, and the three for face spoofing, highlighted in Fig.~\ref{fig:model-fitting}.
The meta-model associated to each cluster, corresponding to a point in the $(\mu_{\alpha}, \sigma_{\alpha})$ plane, can then be used to simulate a given presentation attack scenario, involving the corresponding trait and fake fabrication technique(s), as explained in the next sections.

\begin{table}[tb]
\begin{center}
\caption{Attack scenarios for fingerprint and faces, and their parameters.}
\label{tab:model-parameters}
\vspace{-10pt}
\begin{tabular}{|p{1.2cm}|p{0.4cm}|p{0.4cm}|p{0.8cm}|p{3cm}|}
\hline
\emph{Fing.} & $\mu_\alpha$ & $\sigma_\alpha$ & \emph{risk} & \emph{Dataset(s)}  \\
\hline
Low risk & 0.08 & 0.09 & 0.28\% & LivDet09-Silicone; LivDet11-Alginate\\
Med. risk & 0.23 & 0.20 & 12.33\%  & LivDet11-Gelatin\\
High risk & 0.40 & 0.26  & 35.67\% & LivDet11-Silicone; LivDet11-Latex\\
\hline
\hline
\emph{Face} & $\mu_\alpha$ & $\sigma_\alpha$ & \emph{risk}  & \emph{Dataset(s)} \\
\hline
Low risk & 0.38 & 0.03  & 0.01\% & Personal Photo Attack; Mask Attack\\
Med. risk & 0.78 & 0.19  & 89.83\% & Print Attack\\
High risk & 0.91 & 0.11  & 98.75\% & Photo Attack\\
\hline
\end{tabular}
\end{center}
\vspace{-10pt}
\end{table}

Fig.~\ref{fig:beta-attack-scenarios} shows the Beta distributions associated to the attack scenarios of Fig.~\ref{fig:model-fitting} (using the cluster centroids), and the corresponding values of Eq.~\eqref{eq:risk-level}.
It can be seen that, for each considered trait, the Beta distribution of the different attack scenarios are characterized by considerably different values of the attack impact.
Accordingly, we can label the above scenarios as ``low'', ``medium'' and ``high'' impact. In Table~\ref{tab:model-parameters} we report the corresponding values of $(\mu_{\alpha}, \sigma_{\alpha})$, together with the attack techniques associated to each scenario.
This taxonomy may be clearly revised in the future, if novel attack scenarios emerge from new empirical evidences.

To sum up, our meta-analysis does not only provide a clear picture of current fingerprint and face spoofing attacks, but also the first quantitative characterization of their \emph{impact}.

\vspace{-5pt}
\section{Data Modeling for Multibiometric Systems under Presentation Attacks}
\label{sect:data-model}

We define here a data model for multibiometric systems to account for different presentation attacks against each matcher, and revise the metrics used for evaluating the performance of such systems accordingly. This model will be exploited in the rest of the paper to define our security evaluation procedure, and to design secure fusion rules.

In the following, uppercase and lowercase letters respectively denote random variables (r.v.) and  their values. We denote with $Y \in \{ \rm G, \rm I \}$  the r.v.~representing an identity claim made by either a genuine user (G) or an impostor (I), and with $\mathbf A =(A_{1}, \ldots, A_{K}) \in \set A = \prod_{i=1}^{K} \{0, \ldots, u_{i}\}$,  the r.v.~denoting whether the $i^{\rm th}$ matcher is under attack ($A_i \neq 0$) or not ($A_i = 0$), assuming that $u_{i} > 0$ different presentation attacks are possible against the $i^{\rm th}$ matcher.

For instance, in Sect.~\ref{sect:attack-scenario-definition}, we found three representative attack scenarios for fingerprint and face. To model them, the corresponding $A_{i}$ should take values in $\{ 0, 1, 2, 3 \}$ (\ie, $u_{i}=3$), respectively denoting the no-spoof scenario ($A_i = 0$), and the low-, medium- and high-impact scenarios ($A_i = 1, 2, 3$).

\begin{figure}[t]
\begin{center}
\includegraphics[width=0.22\textwidth]{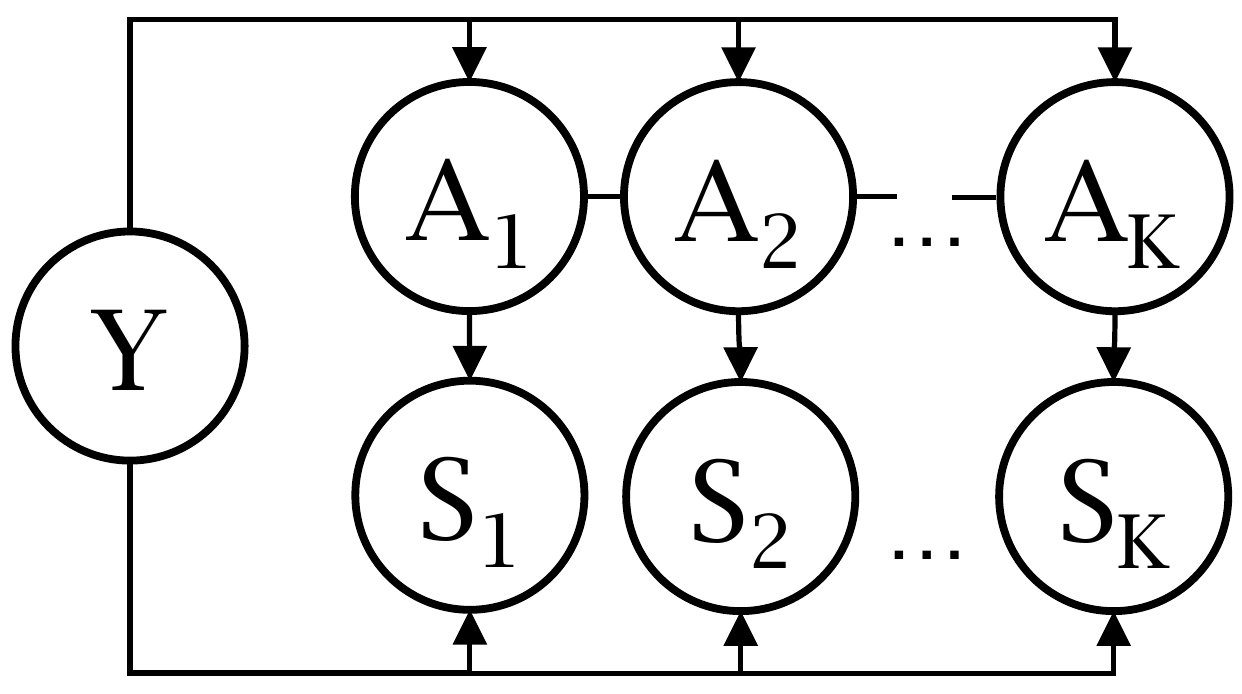}
\vspace{-8pt}
\caption{A Bayesian network equivalent to Eq.~\eqref{eq:smc-model}.}
\vspace{-18pt}
\label{fig:smc-model}
\end{center}
\end{figure}

Assuming that the matching scores $\mathbf S = (S_1, \ldots, S_K) \in \mathbb R^K$ are independent from each other, given $Y$, and that each $A_i$ only influences the corresponding  $S_{i}$, we can write the class-conditional score distributions by marginalizing over all possible values of $\vct A=\vct a \in \set A$, as: 
\begin{align}
p(\mathbf S | Y)  =  \sum_{\mathbf a \in \set A} p(\mathbf a | Y) \prod_{i=1}^K p(S_i | a_i, Y) \enspace .
 \label{eq:smc-model}
\end{align}
Note that the \emph{attack} variables $A_i$ are not assumed to be independent, given $Y$, \ie, $p(\mathbf A | Y) \neq \prod_{i=1}^K p(A_i | Y)$. 
This model corresponds to the Bayesian network of Fig.~\ref{fig:smc-model}.

Since it is unlikely that genuine users use fake traits to access the system, we can reasonably set $p(\mathbf A = \{0\}^K | Y= {\rm G}) = 1$. Thus, the genuine distribution consists of a single component, \ie, $p(\vct S | Y={\rm G}) = p(\vct S | \vct A = \{0\}^{K}, Y={\rm G})$.
The impostor distribution $p(\vct S | Y={\rm I})$ is instead modeled as a mixture of $ |\set A|  = \prod_{i=1}^{K} (u_{i}+1)$ different components, including the distribution $p(\vct S | \vct A = \{0\}^{K}, Y={\rm I})$ of zero-effort impostors, and the distributions $p(\vct S | \vct A = \vct a, Y={\rm I})$, for $\vct a \in \set A \setminus \{0\}^{K}$, associated to different combinations of attacked matchers and presentation attacks.

Accordingly, for a given fusion rule $f(\cdot)$ and acceptance threshold $t$, the metrics used to evaluate the performance of a multibiometric system can be defined as:
\begin{align}
{\rm FRR} &= p(f(\mathbf S) < t | Y={\rm G}) \,, \\
{\rm FAR} &= p(f(\mathbf S) \geq t |  \mathbf A = \{0\}^K, Y={\rm I}) \, , \\
{\rm SFAR}_{\vct a} &= p(f(\mathbf S) \geq t | \mathbf A = \vct a, Y={\rm I}), \vct a \in \set A \setminus \{0\}^{K} \,,  
\end{align}
where ${\rm SFAR}_{\vct a}$ denotes the ${\rm SFAR}$ associated to a specific combination $\vct a \neq \{0\}^{K}$ of attacked matchers and spoofing attacks (which are $|\set A|-1$ in total).
Further, it is not difficult to see that the so-called Global FAR (${\rm GFAR}$)~\cite{chingovska14-spoof-chapter,hadid15-spmag} attained in the presence of a mixture of zero-effort and spoof impostors can be directly computed as a convex linear combination of ${\rm FAR}$ and ${\rm SFAR}$ using Eq.~\eqref{eq:smc-model}, as:
\begin{align}
\nonumber {\rm GFAR} &= p(f(\mathbf S) > t | {\rm I}) 
= \sum_{\mathbf a \in \set A}  p(\mathbf a | {\rm I})  \int_{t}^{\infty} p(f(\mathbf S) | \vct a, {\rm I} ) df(\mathbf S)\\
&=  \pi_{\vct 0} \, {\rm FAR} +  \sum_{\vct a \in \set A \setminus \{0\}^{K}} \pi_{\vct a} \, {\rm SFAR}_{\vct a} \enspace , \label{eq:GFAR} 
\end{align}
where, for notational convenience, we set $\pi_{\vct 0}=p(\vct A=\{0\}^{K} | {\rm I})$ and $\pi_{\vct a}=p(\vct A = \vct a | {\rm I})$, for $\vct a \neq \{0\}^{K}$.
To our knowledge, this is the first model highlighting a clear connection between the aforementioned performance metrics and the distribution of zero-effort and spoof impostors.

\vspace{-5pt}
\section{Security Evaluation of Multibiometric Systems under Presentation Attacks} 
\label{sect:method}

In this section we describe our security evaluation procedure, and show how it can be exploited also for selecting the fusion rule and/or its parameters.

\textbf{Security evaluation}. 
The procedure we propose is \emph{empirical}, as in \cite{rodrigues09,rodrigues10,johnson10}: the SFAR under a simulated presentation attack is evaluated by replacing the available zero-effort impostor scores coming from the attacked matchers with fictitious fake scores sampled from our meta-model.
More precisely, consider a multibiometric system made up of $K$ matchers, and an available set of matching scores $\mathcal D = \{s^{1}_{j}, \ldots, s^{K}_{j}, y_{j}\}_{j=1}^{n}$, with $y_{j} \in \{{\rm G}, {\rm I} \}$.
For a point estimate of the SFAR under a \emph{single} attack against a subset of matchers (\eg, one of the known attacks), one should first define the combination $\vct a \in \set A \setminus \{0\}^{K}$ of attacked matchers and attack scenarios, \ie, the values of the meta-parameters $(\mu_\alpha, \sigma_\alpha)$ for each such matcher. Then:
\begin{itemize}
\item[($i$)] for each matcher $i=1,\ldots,K$, if $a_{i} \neq 0$, set $(\mu^{i}_\alpha, \sigma^{i}_\alpha)$ according to the desired scenario; 
\item[($ii$)] set $\mathcal D^{\prime}=\{s^{\prime 1}_{j}, \ldots, s^{\prime K}_{j}, y^{\prime}_{j}\}_{j=1}^{n}$ equal to $\mathcal D$;
\item[($iii$)] for each $(i,j)$, if $y_{j}={\rm I}$ and $a_{i} \neq 0$, set $s^{\prime i}_{j}$ according to Eq.~\eqref{eq:alpha-model}, with $\alpha$ drawn from $p(\alpha | \mu_{\alpha}^{i},\sigma_{\alpha}^{i})$, and  $S^{\rm G}$ and $S^{\rm I}$ sampled from $\{s^{i}_{j}\}_{j | y_{j}={\rm G}}$ and $\{s^{i}_{j}\}_{j | y_{j}={\rm I}}$, \ie, the genuine and impostor scores of the $i^{\rm th}$ matcher; 
\item[($iv$)] evaluate ${\rm SFAR}_{\vct a}$ empirically using $\mathcal D^{\prime}$.
\end{itemize}
The resulting ${\rm GFAR}$ can then be evaluated by Eq.~\eqref{eq:GFAR} (where the summation reduces to the single term $\vct a$), after estimating the ${\rm FAR}$ through the standard procedure and hypothesizing the values of $\pi_{\vct 0}$ and $\pi_{\vct a}$.

To evaluate the ${\rm GFAR}$ of Eq.~\eqref{eq:GFAR} under different combinations $\vct a \neq \{0\}^{K}$ of attacked matchers and scenarios, the above steps can be repeated for each of them.

\textbf{Uncertainty analysis for security evaluation.} To account for both known and unknown presentation attacks, through an uncertainty analysis, the above procedure has to be carried out by sampling a large number of attack scenarios $(\mu_{\alpha},\sigma_{\alpha})$ from the feasible region of Fig.~\ref{fig:model-fitting}, besides the known attacks of interest. For instance, a uniform sampling can be used, as we will show in Sect.~\ref{sect:exp}.
In particular, it is convenient to sort the attack scenarios of each matcher for increasing values of \emph{attack impact} (Eq.~\ref{eq:risk-level}), such that higher $a_{i}$ values correspond to a higher attack impact. This allows the ${\rm SFAR}_{\vct a}$ to be evaluated as a function of the attack impact on each matcher, highlighting its variability around the point estimates corresponding to known attacks, and which matchers the fusion rule is most sensitive to.
Confidence bands can be used to represent the variability of the ${\rm SFAR}_{\vct a}$ as the attack impact varies, as commonly done in statistical data analysis to represent the uncertainty on the estimate of a curve or function based on limited or noisy data.
Examples are given in the plots of Fig.~\ref{fig:fing-face}, where the yellow and purple bands represent the uncertainty on the impact of different attacks on the ${\rm SFAR}$. As we will show in Sect.~\ref{sect:exp}, even the ${\rm SFAR}$ associated to never-before-seen attacks can fall within these confidence bands, highlighting that our approach can also predict the impact of unknown attacks on the system.

\textbf{Fusion rule and parameter selection}. Our security evaluation procedure can be also exploited to help system designers selecting an appropriate fusion rule (or tuning its parameters), taking into account a set of potential attack scenarios.
Assume that the designer has identified a number of relevant attack scenarios $\vct a$ of interest, and would like to choose among $p \geq 1$ fusion rules $F = \{f_1,\ldots,f_p\}$, and/or tuning their parameter vectors $\Theta=\{\theta_1,\ldots,\theta_p\}$, to attain a suitable trade-off between the performance in the absence of attacks (defined by ${\rm FRR}$ and ${\rm FAR}$ values) and the one under the relevant attack scenarios, defined by the corresponding values of ${\rm SFAR}_{\vct a}$, estimated using the above procedure.
While in the absence of attack application requirements can be expressed in terms of a trade-off between ${\rm FRR}$ and ${\rm FAR}$, in the presence of spoofing the desired trade-off should also account for the ${\rm SFAR}$ under distinct, potential attack scenarios~\cite{rodrigues09,rodrigues10,johnson10,fumera14-spoof-chapter,chingovska14-spoof-chapter,hadid15-spmag}.
This can be expressed, for instance, by minimizing a desired ${\rm GFAR}$ expression (Eq.~\ref{eq:GFAR}), while keeping the ${\rm FRR}$ below a maximum admissible value ${\rm FRR}_{\max}$:
\begin{eqnarray} 
\label{eq:criterion1}
\min_{F,\Theta} & & {\rm GFAR} =
\pi_{\vct 0} \, {\rm FAR} +  \sum_{\vct a \in \set A \setminus \{0\}^{K}} \pi_{\vct a} \, {\rm SFAR}_{\vct a}  \, , \\
\label{eq:criterion2}
{\rm s.t. } && {\rm FRR} \leq {\rm FRR}_{\max} \, . 
\end{eqnarray}
The value of ${\rm FRR}_{\max}$ and the priors $\pi_{\vct 0}$ and $\pi_{\vct a}$ have to be carefully chosen by the system designer, depending on the application at hand, and on the attack scenarios that are considered more relevant.
In Sect.~\ref{sect:exp} we will show an example of how to exploit the proposed procedures to assess the security of a bimodal system against fingerprint and face presentation attacks, and to select a suitable fusion rule.

\vspace{-5pt}
\section{Design of Secure Fusion Rules}
\label{sect:secure-fusion}

The secure score-level fusion rules proposed so far are based on explicitly modeling presentation attacks against each matcher as part of the impostor distribution, using the scenario defined by Eq.~\eqref{eq:assumption-rodrigues}~\cite{rodrigues09,rodrigues10,johnson10}. 
However, as discussed in Sect.~\ref{sect:background}, this may cause such rules to exhibit a too pessimistic trade-off between the performance in the absence of spoofing and that under attacks that are not properly represented by Eq.~\eqref{eq:assumption-rodrigues}.
In this section, we discuss how to overcome these limitations using our meta-model (Sect.~\ref{sect:meta-analysis}).

We first show how previously-proposed secure fusion rules can be interpreted according to the data model of Sect.~\ref{sect:data-model} (Sect.~\ref{sect:secure-fusion-old}), and how this model can be also exploited to train secure fusion rules based on discriminative classifiers (Sect.~\ref{sect:discriminative-approach}). Then, we discuss how our meta-model of $p(S^{\rm F})$ and the attack scenarios defined in Sect.~\ref{sect:attack-scenario-definition} can be exploited to design spoofing-aware score-level fusion rules that can achieve a better trade-off in terms of ${\rm FRR}$, ${\rm FAR}$ and ${\rm SFAR}$, on a wider set of attack scenarios characterized by different levels of \emph{attack impact}. In particular, two secure score-level fusion rules are considered, respectively relying on a generative and a discriminative approach (Sect.~\ref{sect:secure-fusion-new}).

\vspace{-5pt}
\subsection{Previously-proposed Secure Fusion Rules}
\label{sect:secure-fusion-old}

In~\cite{biggio11-smc,rodrigues09} spoofing-aware score-level fusion rules were proposed, as variants of the well-known LLR rule~\cite{nandakumar08}:
\begin{equation}
f(\mathbf{s}) = p(\mathbf{s}|Y={\rm G}) / p(\mathbf{s}|Y={\rm I}) \enspace .
\label{eq:LLR}
\end{equation}
Both exploit an estimate of $p(\mathbf S | Y={\rm I})$ incorporating knowledge of \emph{potential} presentation attacks that may be incurred during operation, and that are not included in the training data (as our model of Eq.~\ref{eq:smc-model}).
Therefore, while the genuine and the zero-effort impostor distributions can be estimated from the corresponding matching scores in the training data, specific assumptions have to be made on the remaining components of the mixture $p(\mathbf S | Y={\rm I})$ of zero-effort and spoof impostors. 
In our model, they include the priors $p(\mathbf a | Y={\rm I})$ and the fake score distributions $p(S_i | a_{i}, Y={\rm I})$, for $a_{i}=1,\ldots,u_{i}$, and $i=1,\ldots,K$.
Both rules assume only a possible kind of attack against each matcher. This can be easily accounted for in our model of Eq.~\eqref{eq:smc-model} by setting $\vct A \in \{0,1\}^{K}$, \ie, $u_{i}=1$ for $i=1,\ldots,K$.

\textbf{Extended LLR}~\cite{rodrigues09}. This rule is based on a seemingly more complex expression of $p(\mathbf S | Y={\rm I})$ than Eq.~ \eqref{eq:smc-model}, as it includes the probability of \emph{attempting} a presentation attack against each matcher (represented by the r.v.~$\mathbf T \in \{0,1\}^{K}$), and the probability of each attempt being \emph{successful} (represented by the r.v.~$\mathbf F \in \{0,1\}^{K}$).
For each matcher, only if an attack is attempted and successful (\ie, $T_{i}=1$ and $F_{i}=1$), then the corresponding score follows a distribution different from that of zero-effort impostors.
The expression of $p(\mathbf S | Y={\rm I})$ becomes however equivalent to Eq.~\eqref{eq:smc-model}, if we set $A_i = F_i$, and marginalize over $T_i$ (\cf{} Eq.~\ref{eq:smc-model} with Eq.~5 in \cite{rodrigues09}). The prior distribution can be indeed written as:
\begin{equation}
\label{eq:rodrigues}
p(\mathbf A | Y={\rm I})  = \sum_{\mathbf t \in \{0,1\}^{K}} p(\mathbf t | Y={\rm I}) \prod_{i=1}^K p(A_i | t_i) \enspace ,
\end{equation}
while the fake score distributions $p(S_{i} | F_{i}, Y={\rm I})$ in \cite{rodrigues09} are clearly equivalent to our $p(S_{i} | A_{i}, Y={\rm I})$.

In \cite{rodrigues09} the probability of \emph{attempting} a presentation attack against any of the $2^K-1$ combinations of matchers $p(\mathbf T \neq \{0\}^{K}  | Y={\rm I})$ was set to $r / (2^K-1)$. Thus, the probability of zero-effort impostor attempts $p(\mathbf T = \{0\}^{K}  | Y={\rm I})$ was set to $1-r$, being $r$ a parameter.\footnote{To avoid confusion, we use $r$ instead of $\alpha$ as in \cite{rodrigues09}.}
The probability of an attempted spoof \emph{failing}, $p(A_i=0 | T_i = 1)$, was denoted as $c_i$, and referred to as the ``level of security'' of the $i^{\rm th}$ matcher.
Clearly, as an attack can not be successful if it has not been attempted, $p(A_i=1 | T_i = 0)=0$.
The resulting expression of $p(\mathbf S | Y={\rm I})$ therefore depends on the parameters $r$ and $c_i$. Setting their values amounts to defining the distribution $p(\mathbf A|Y={\rm I})$ in Eq.~\eqref{eq:smc-model} (see Eqs.~\ref{eq:smc-model} and~\ref{eq:rodrigues}); \eg, for a bimodal system ($K=2$), one obtains:
\begin{align}
\label{eq:spoof-prior-1}
p(A_1=1, A_2=0 | {\rm I}) = \frac{r}{3}(1-c_1)(1+c_2) ,\\
\label{eq:spoof-prior-2}
p(A_1=0, A_2=1 | {\rm I}) = \frac{r}{3}(1+c_1)(1-c_2) ,\\
\label{eq:spoof-prior-3}
p(A_1=1, A_2=1 | {\rm I}) = \frac{r}{3}(1-c_1)(1-c_2),\\
\label{eq:spoof-prior-4}
p(A_1=0, A_2=0 | {\rm I}) = \frac{r}{3}(c_1+c_2+c_1c_2) + 1-r .
\end{align}
Notably, if we assume $\vct A \in \{0,1\}^{K}$ and the scenario of Eq.~\eqref{eq:assumption-rodrigues}, the distribution $p(\mathbf S |Y={\rm I})$
described by the models of Eqs.~\eqref{eq:smc-model} and~\eqref{eq:rodrigues}
becomes identical, as well as the corresponding LLR-based secure fusion rules given by Eq.~\eqref{eq:LLR}.

\textbf{Uniform LLR}~\cite{biggio11-smc}. We proposed this robust version of the LLR in previous work, for a broader class of applications in computer security.
It is based on modeling the impostor distribution according to Eq.~\eqref{eq:smc-model}, with $\vct A \in \{0,1\}^{K}$. However, we considered the case when no specific knowledge on the distribution of potential attacks is available to the designer; accordingly, we \emph{agnostically} assumed a uniform distribution for modeling $p(S_i | A_i = 1, Y={\rm I})$.

\subsection{Discriminative Approaches}
\label{sect:discriminative-approach}

The aforementioned secure fusion rules are based on a \emph{generative} model of the data distribution.
However, Eq.~\eqref{eq:smc-model}
can be also exploited to develop secure rules based on \emph{discriminative} classifiers such as Support Vector Machines (SVMs) and Neural Networks (NNs)~\cite{bishop:prml:book:2007}.
To this end, one may train the fusion rule after resampling the available zero-effort impostor scores according to Eq.~\eqref{eq:smc-model} as follows (see also \cite[Sect.~8.1.2]{bishop:prml:book:2007}).
First, define $p(\mathbf S| {\rm I})$ according to Eq.~\eqref{eq:smc-model}, \ie, define $p(\mathbf A | Y={\rm I})$ and $p(S_i | A_{i}, Y={\rm I})$ (for $A_{i}=1,\ldots,u_{i}$, and $i=1,\ldots,K$).
As for the security evaluation procedure defined in Sect.~\ref{sect:method}, let us denote the available set of scores as $\mathcal D = \{s^{1}_{j}, \ldots, s^{K}_{j}, y_{j}\}_{j=1}^{n}$, with $y_{j} \in \{{\rm G}, {\rm I} \}$. For each $j | y_{j}={\rm I}$,  draw a value of $\vct a$ from $p(\mathbf A | {\rm I})$.
For each $a_{i} \in \vct a$, if $a_{i}\neq 0$, replace the corresponding $s_{i}$ with a sample from the hypothesized fake score distribution $p(S_{i}| a_{i},{\rm I})$, otherwise leave $s_{i}$ unmodified (\ie, sample from the empirical distribution of zero-effort impostors). 

Despite their simplicity, discriminative approaches have not been widely considered to design secure fusion rules.
We will show how to exploit them to this end in Sect.~\ref{sect:exp}.

\vspace{-6pt}
\subsection{Secure Fusion Rules based on our Meta-model} 
\label{sect:secure-fusion-new}

We showed that our meta-model (Eq.~\ref{eq:alpha-model}) can be exploited in the design of secure fusion rules to simulate the fake score distribution $p(S_i | A_i \neq 0, Y={\rm I})$ at the output of each attacked matcher, both in generative and discriminative approaches.
To overcome the limitations of secure fusion rules based on Eq.~\eqref{eq:assumption-rodrigues}, the aforementioned distribution can be hypothesized by selecting a suitable combination of attack scenarios, depending on the given application and desired level of security.
The corresponding parameters $(\mu_{\alpha}, \sigma_{\alpha})$ can be selected either among those defined in Sect.~\ref{sect:attack-scenario-definition} (Table~\ref{tab:model-parameters}) for faces and fingerprints, or through cross validation, to properly tune the trade-off among ${\rm FRR}$, ${\rm FAR}$ and ${\rm SFAR}$ (or ${\rm GFAR}$) under a wider class of presentation attacks. 

In the following, we give two examples of how novel secure fusion rules can be defined.
Different choices are possible, depending on the the selected fusion scheme (\eg, LLR, SVM, NN), and on the trade-off between the performance in the absence of attack and the security level that one aims to achieve. In our data model (Eq.~\ref{eq:smc-model}), this influences the choice of each prior $p(\vct A | Y={\rm I})$ and of the corresponding attack scenarios (\ie, the parameters $\mu_\alpha$ and $\sigma_\alpha$).
We consider here an application setting demanding for a high level of security, \eg, an access control system for banking, and we thus only consider the worst-case available scenarios involving \emph{high-impact} presentation attacks (see Table~\ref{tab:model-parameters}).

\textbf{$\alpha$-LLR.} This is another variant of the generative LLR rule (Eq.~\ref{eq:LLR}), %
in which $\vct A =\{0,1\}^{K}$ and the distributions $p(S_i | A_i = 1, Y={\rm I})$ are simulated according to the \emph{high-impact} attack scenario. The prior distribution $p(\vct A | Y={\rm I})$ should be hypothesized based on the specific application setting, as suggested for the Extended LLR~\cite{rodrigues09}.

\textbf{$\alpha$-SVM-RBF.} This rule is instead based on a discriminative approach. It consists of learning an SVM with the Radial Basis Function (RBF) kernel on a modified training set that includes simulated presentation attacks, as discussed in Sect.~\ref{sect:discriminative-approach}:
the available impostor scores are replaced with a number of matching scores sampled from the hypothesized $p(\vct S | Y={\rm I})$.
As for the $\alpha$-LLR, $p(\vct A | Y={\rm I})$ is a parameter, $\vct A =\{0,1\}^{K}$ and $p(S_i | A_i = 1, Y={\rm I})$ is simulated according to the \emph{high-impact} attack scenario.

Besides $p(\vct A | Y={\rm I})$, the other parameters to be tuned are the SVM regularization parameter $C$ and the parameter $\gamma$ of the RBF kernel, given by $k(\vct s, \vct s_{i})=\exp \left ( -\gamma ||\vct s - \vct s_{i} ||^{2}\right )$, where $\vct s$ and $\vct s_{i}$ denote the input and the $i$-th training score vectors~(see, \eg, \cite{bishop:prml:book:2007}).

In conclusion, it is worth remarking that each secure fusion rule makes specific assumptions on the mixture of zero-effort and spoof impostors $p(\vct S | Y={\rm I})$ (Eq.~\ref{eq:smc-model}), in terms of the prior $p(\vct A | Y={\rm I})$ and of the fake score distributions $p(S_{i} | A_{i} = a_{i}, Y={\rm I})$, for $i=1,\ldots,K$ and $a_{i} \in \{1, \ldots, u_{i}\}$.
It is thus clear that each rule will achieve an optimal trade-off between ${\rm FRR}$ and ${\rm GFAR}$ (Eq.~\ref{eq:GFAR}) only when the ${\rm GFAR}$ is obtained under the same hypothesized model of $p(\vct S | Y={\rm I})$ (Eq.~\ref{eq:smc-model}).
This also holds for the secure fusion rules proposed in this section. Nevertheless, the data model of Sect.~\ref{sect:data-model} along with the meta-model of Sect.~\ref{sect:meta-analysis} can be exploited to implement secure fusion rules that are optimal according to any other choice of $p(\vct S | Y={\rm I})$, giving us much clearer guidelines to design secure score-level fusion rules, especially if compared to previous work~\cite{rodrigues09,rodrigues10,johnson10}.

\vspace{-3pt}
\section{Experimental Analysis} 
\label{sect:exp}

We report here a case study on a bimodal system combining fingerprint and face, to show how to thoroughly assess its security against presentation attacks, and to select a suitable fusion rule, according to the procedures defined in Sect.~\ref{sect:method}.

For our experiments, we consider a scenario in which the designer ($i$) believes that only presentation attacks against one matcher (either the face or fingerprint one) are likely, and ($ii$) would like to select a fusion rule and/or its parameters to protect the multibiometric system against \emph{worst-case} attacks, assuming that attacks against face and fingerprint are equiprobable, and accepting a maximum ${\rm FRR}$ of 2\%.
According to our approach, and to point ($i$) above, all the available attack scenarios encompassed by our meta-model should be considered against each matcher, to thoroughly evaluate system security. With regard to point ($ii$) above, and according to Sect.~\ref{sect:method}, the system designer should also encode application-specific requirements using Eqs.~\eqref{eq:criterion1}-\eqref{eq:criterion2} to define a proper trade-off among ${\rm FRR}$, ${\rm FAR}$ and ${\rm SFAR}$. Then, the goal defined above can be formalized as:
\begin{align} 
\min \; &  {\rm GFAR}  = \frac{1}{2} {\rm FAR} + \frac{1}{4} \left ({\rm SFAR_{\rm H1}}+ {\rm SFAR_{\rm H2}} \right) \, , \label{eq:exp-1} \\
{\rm s.t. } \; & {\rm FRR} \leq 2\% \, , \label{eq:exp-2}
\end{align}
being ${\rm SFAR_{\rm H1}}$ and ${\rm SFAR_{\rm H2}}$ the ${\rm SFAR}$ attained by the first and the second matcher against the corresponding high-impact attacks, while the other matcher is not under attack.

The above setting can be easily generalized to any other combination of attacks, also targeting multiple biometrics at the same time, or choice of parameters $\pi_{\vct 0}$, $\pi_{\vct a}$ and ${\rm FRR}_{\max}$.

\vspace{-4pt}
\subsection{Experimental Setup}

For these experiments, to validate the predictions of our approach under never-before-seen presentation attacks, we have considered two very recent databases of fake fingerprints and faces that have not been used in the design of our meta-model. They are concisely described below. 

\textbf{LivDet15}~\cite{livdet15}. This database consists of about 16,000 fingerprint images captured by performing multiple acquisitions of \emph{all} fingers of 50 different subjects, with four different optical devices (Biometrika, Green Bit, Digital Persona and Crossmatch). Fingerprint images were acquired in a variety of ways (\eg, wet and dry fingers, high and low pressure) to simulate different operating conditions. Fake fingerprints were fabricated using the cooperative method, with different materials, including Ecoflex, Playdoh, Body Double, silicone and gelatin. 
In our experiments, we use the images acquired with the Crossmatch sensor, and consider each separate finger as a client, yielding 500 distinct clients. We use Bozorth3 as the matching algorithm.

\textbf{CASIA}~\cite{zhang12-casia}. This database consists of 600 videos of alive and fake faces belonging to 50 distinct subjects, captured at high and low resolution, respectively, with a Sony NEX-5 and a standard USB camera.
Three different kinds of presentation attacks are considered: \emph{warped photo}, in which face images are printed on copper paper, and warped to simulate motion; \emph{cut photo}, in which the warped face photo has also eye cuts to simulate blinking; and \emph{video}, in which face images are displayed using a mobile device.
We extract four frames from each video, and rotate and scale face images to have eyes in the same positions. We use the same matcher described in~\cite{tuveri15-iciap}. It accounts for illumination variations as in~\cite{tan10-tip}, and then computes a BSIF descriptor~\cite{kannala12-icpr}. Matching scores are finally computed using the cosine distance.

We exploit these unimodal matching scores to create a \textit{chimerical} dataset, 
by randomly associating face and fingerprint images of \emph{different} clients from the two databases. This is a common practice in biometrics to obtain multimodal datasets~\cite{multiBiometrics}.
The chimerical dataset is then randomly subdivided into five pairs of training and test sets, respectively including 40\% and 60\% of the ``virtual'' clients.\footnote{The clients of a chimerical dataset are usually referred to as ``virtual'' clients, as they do not correspond to a real person or identity.}
The matching scores are normalized in $[0,1]$ using the ${\rm min}$-${\rm max}$ technique~\cite{jain05,multiBiometrics}. Its parameters, and those of the trained fusion rules, are estimated on the training set.
This procedure is repeated five times, each time creating a different set of ``virtual'' clients. 
The results thus refer to the average test set performance on the corresponding twenty-five runs.

We consider the following state-of-the-art score-level fusion rules, including the spoofing-aware fusion rules discussed in Sect.~\ref{sect:secure-fusion-old}, and the two secure fusion rules based on our meta-model, as described in Sect.~\ref{sect:secure-fusion-new}.

\textbf{Sum.} Given $K$ matching scores to be combined $\mathbf{s} = (s_1, \ldots, s_K)$, the sum rule is defined as $f(\mathbf{s}) = \sum_{i=1}^K s_i$. 

\textbf{Product.} The product rule is defined as $f(\mathbf{s}) = \prod_{i=1}^K s_i$.

\textbf{Minimum.} This rule is defined as $f(\mathbf{s}) = \min_{i=1}^K s_i$.\footnote{Note that this rule is equivalent to an ``AND'' fusion rule that classifies a claim as genuine only if \emph{all} the combined matchers output a genuine decision, assuming the same threshold for all matchers.}

\textbf{Linear Discriminant Analysis (LDA).} This is a trained rule, in which the matching scores are linearly combined as $f(\mathbf{s}) = \sum_{i=1}^K w_i s_i + b$. The parameters $w_i$ and $b$ are estimated from the training set by maximizing the Fisher distance between genuine and impostor score distributions~\cite{bishop:prml:book:2007}.

\textbf{Likelihood ratio (LLR).} This is the trained rule given by Eq.~\eqref{eq:LLR}. To estimate the  likelihoods $p(\mathbf{s}|Y)$ of genuine and impostor users, it is often realistically assumed that the $s_i$ are independent given $Y$, \ie, $p(\mathbf{s}|Y) = \prod_{i=1}^K p(s_i|Y)$. Here we make the same assumption, and estimate each component $p(s_i|Y)$ by fitting a Gamma distribution on the corresponding training data, as in \cite{rodrigues09,rodrigues10,johnson10}.

\begin{figure*}[t]
\begin{center}
\includegraphics[width = 0.23\textwidth]{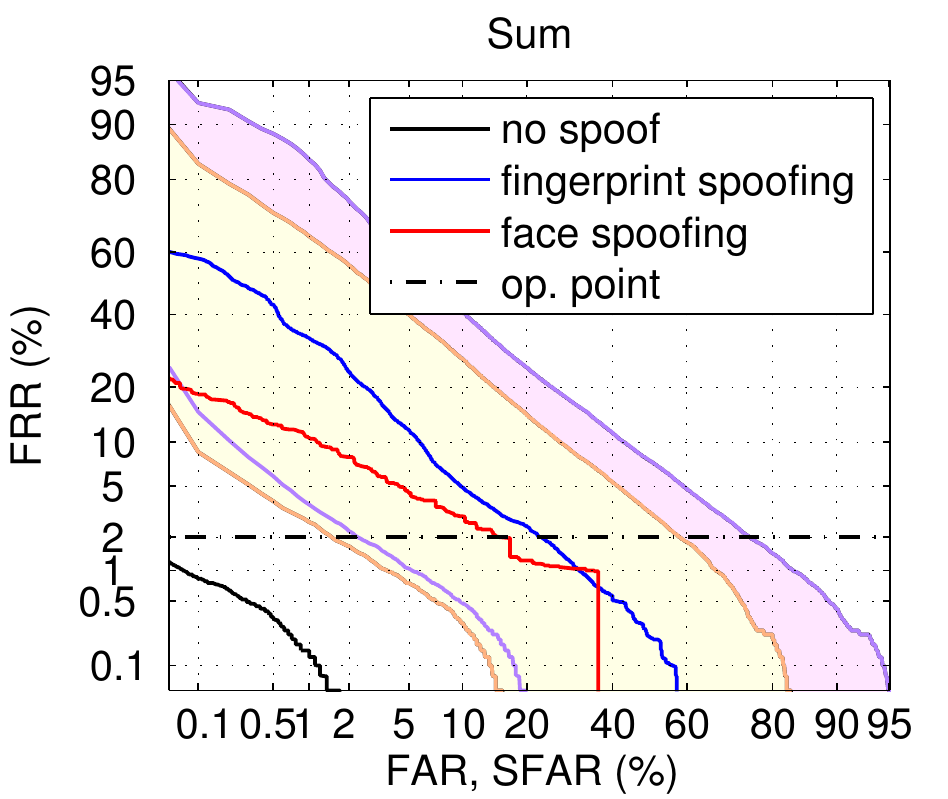}
\includegraphics[width = 0.22\textwidth]{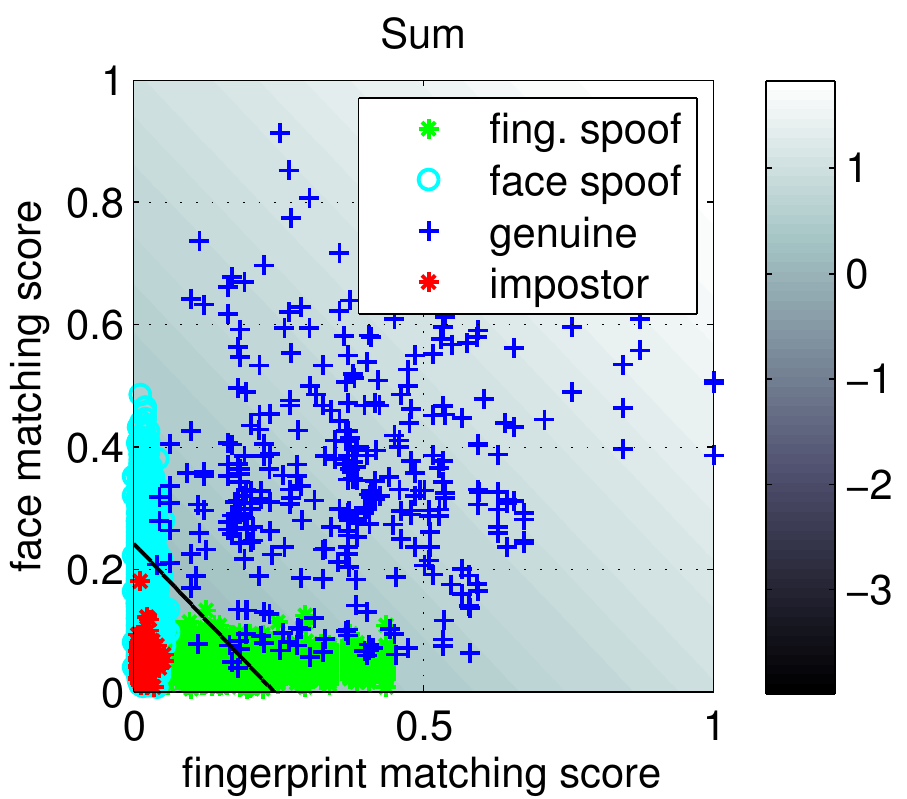}
\includegraphics[width = 0.23\textwidth]{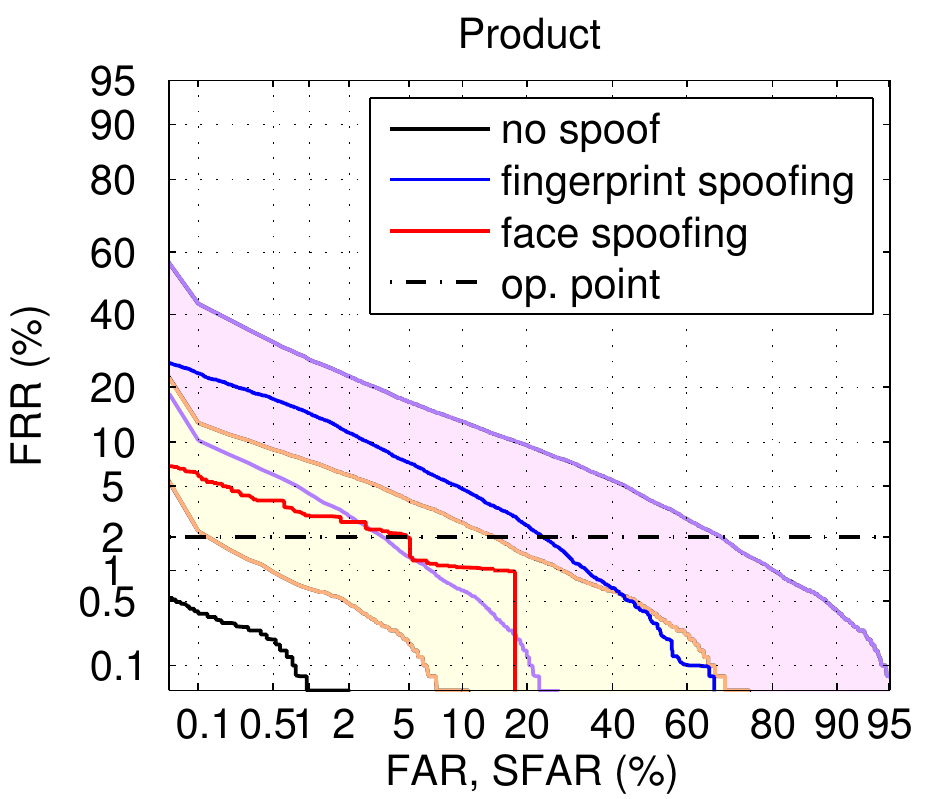}
\includegraphics[width = 0.22\textwidth]{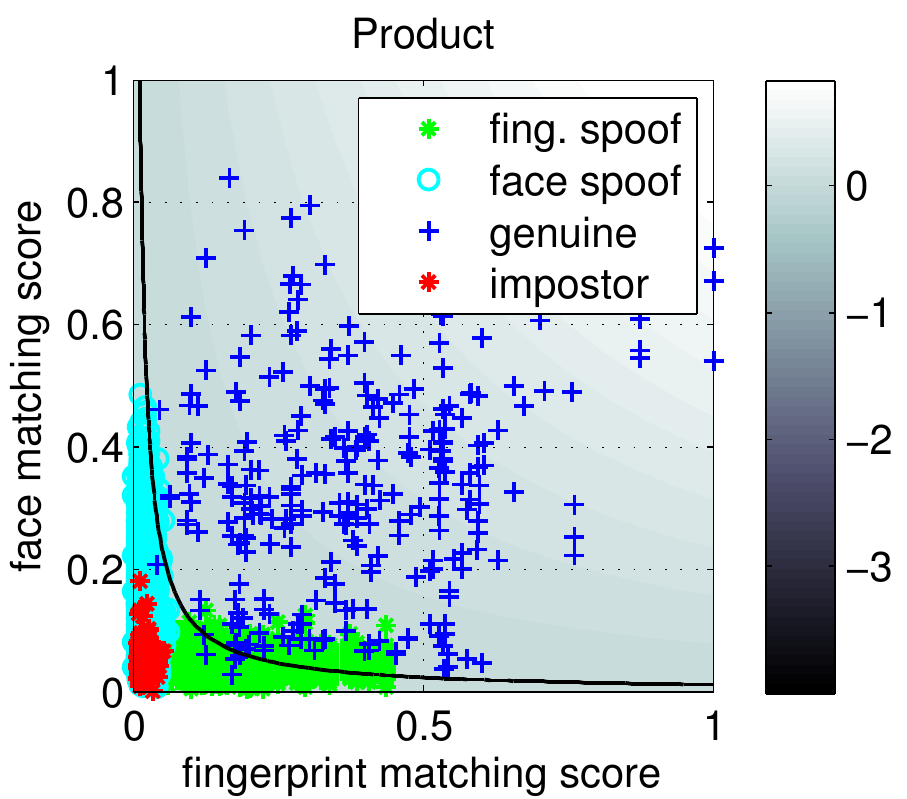}
\includegraphics[width = 0.23\textwidth]{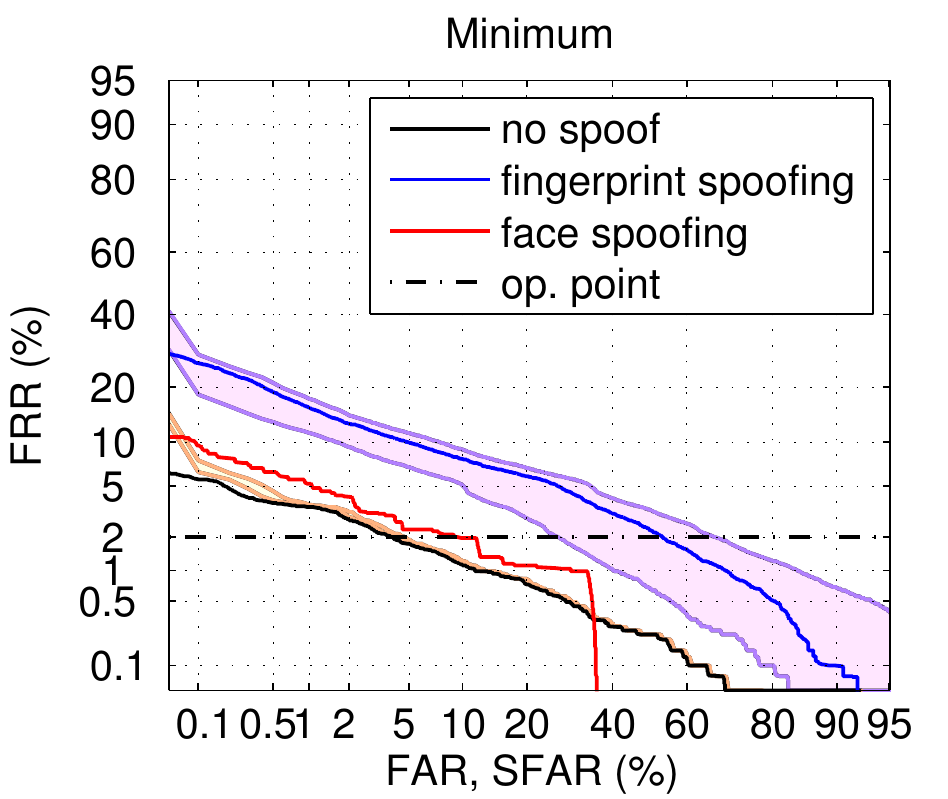}
\includegraphics[width = 0.22\textwidth]{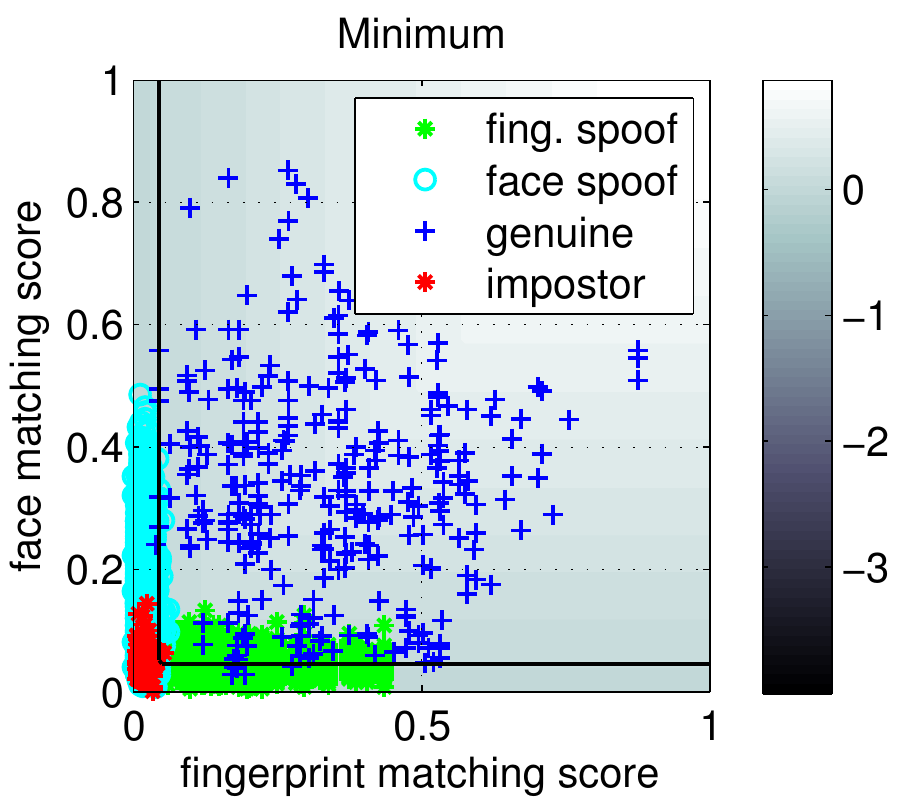}
\includegraphics[width = 0.23\textwidth]{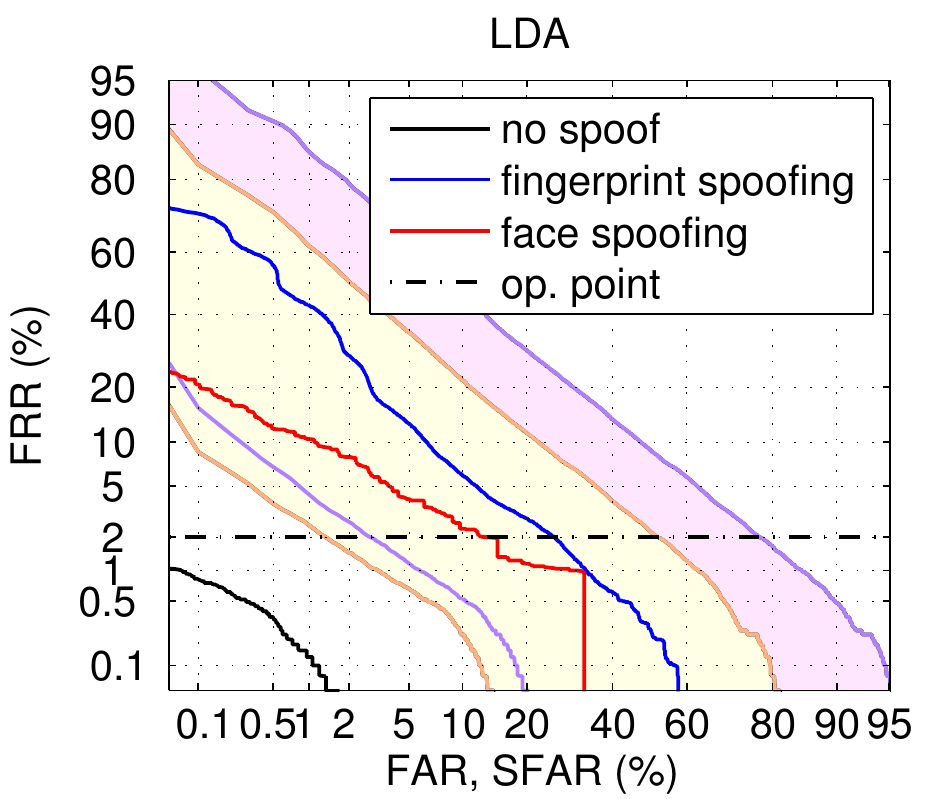}
\includegraphics[width = 0.22\textwidth]{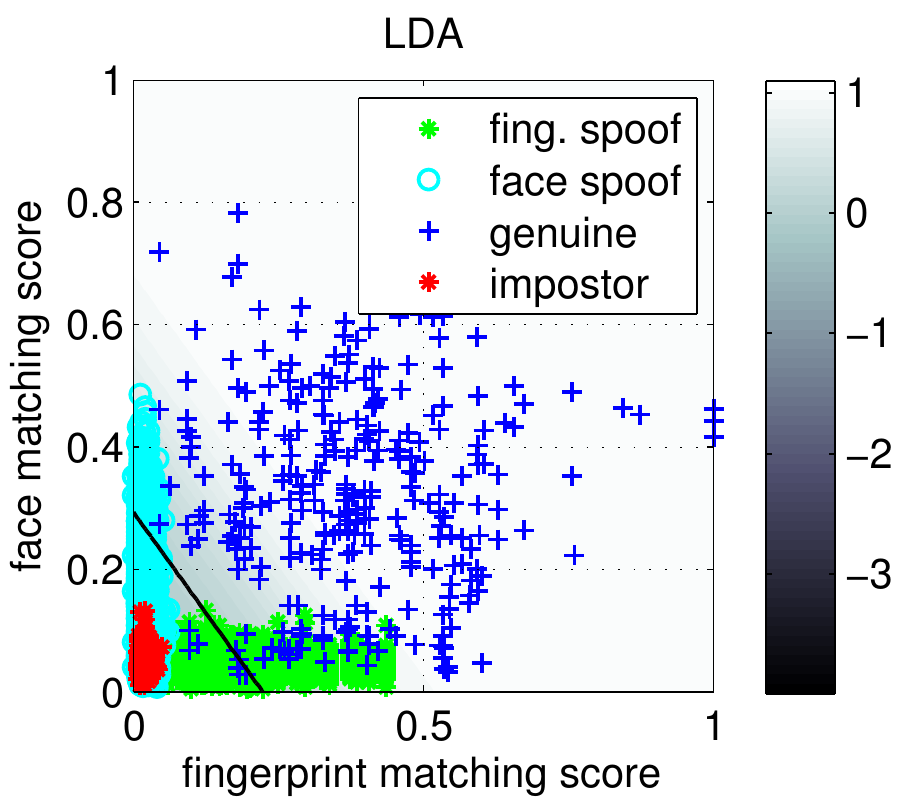}
\includegraphics[width = 0.23\textwidth]{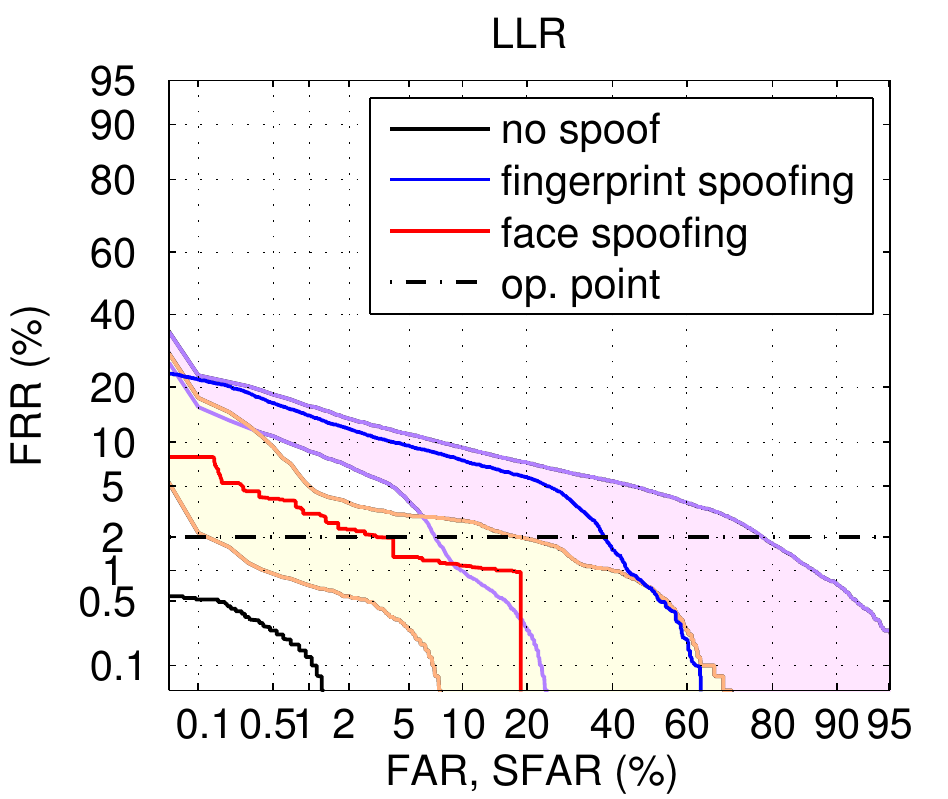}
\includegraphics[width = 0.22\textwidth]{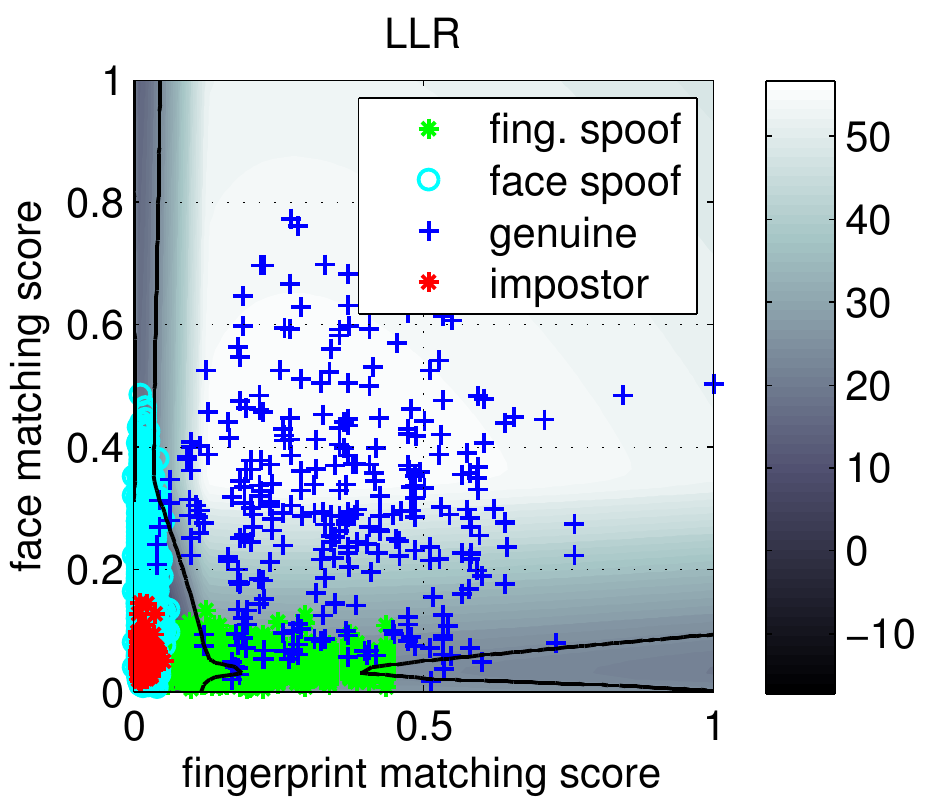}
\includegraphics[width = 0.23\textwidth]{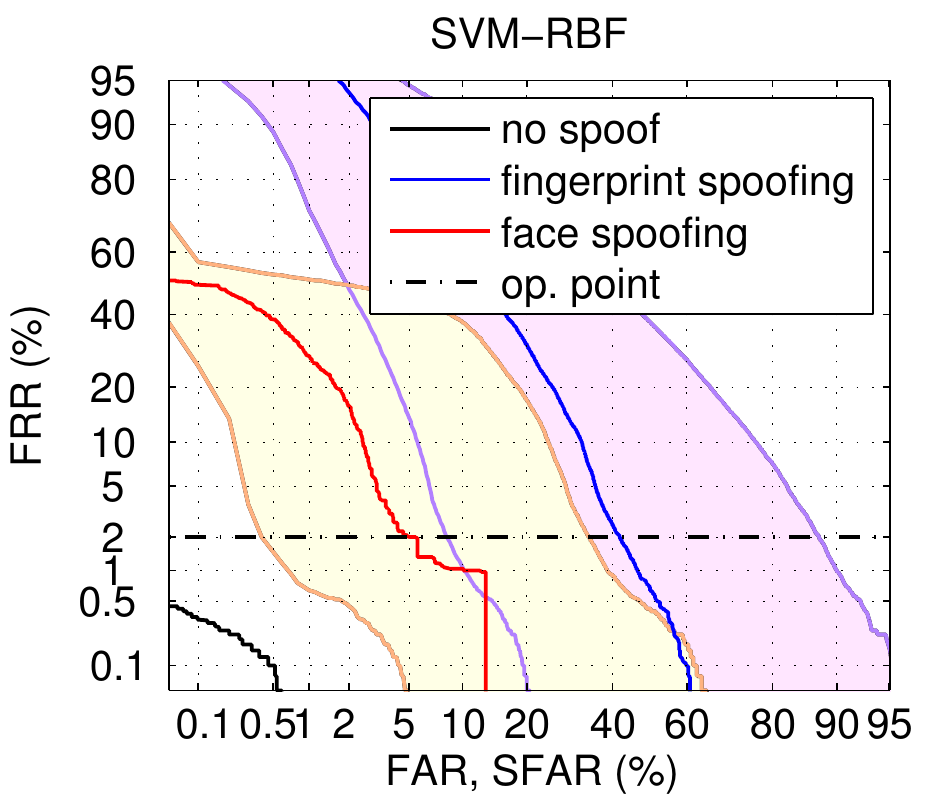}
\includegraphics[width = 0.22\textwidth]{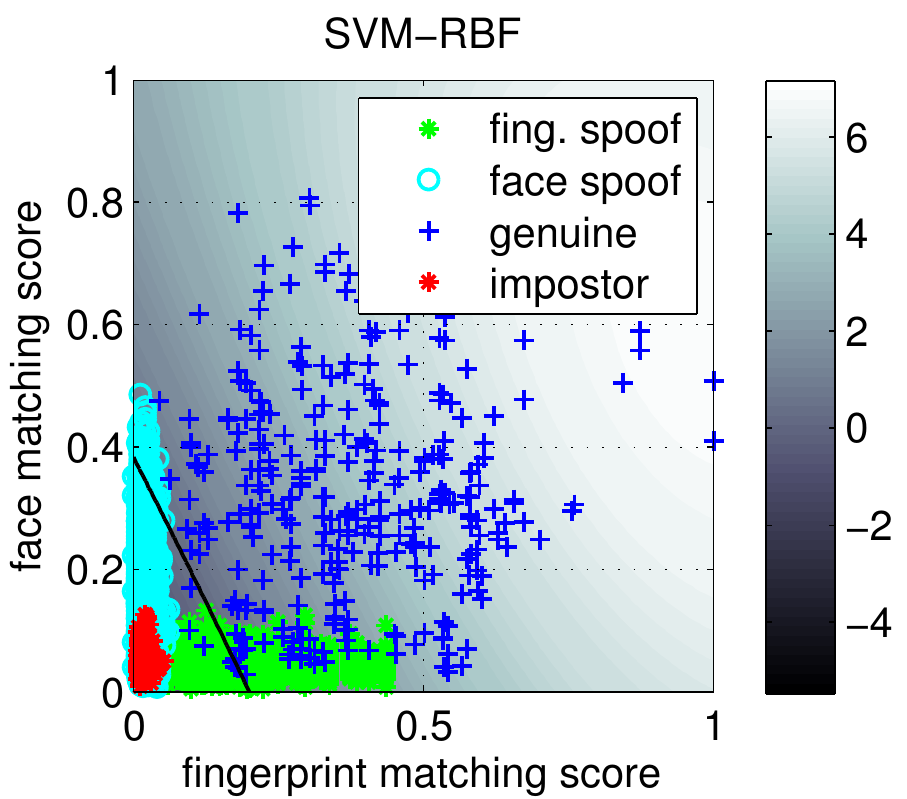}
\includegraphics[width = 0.23\textwidth]{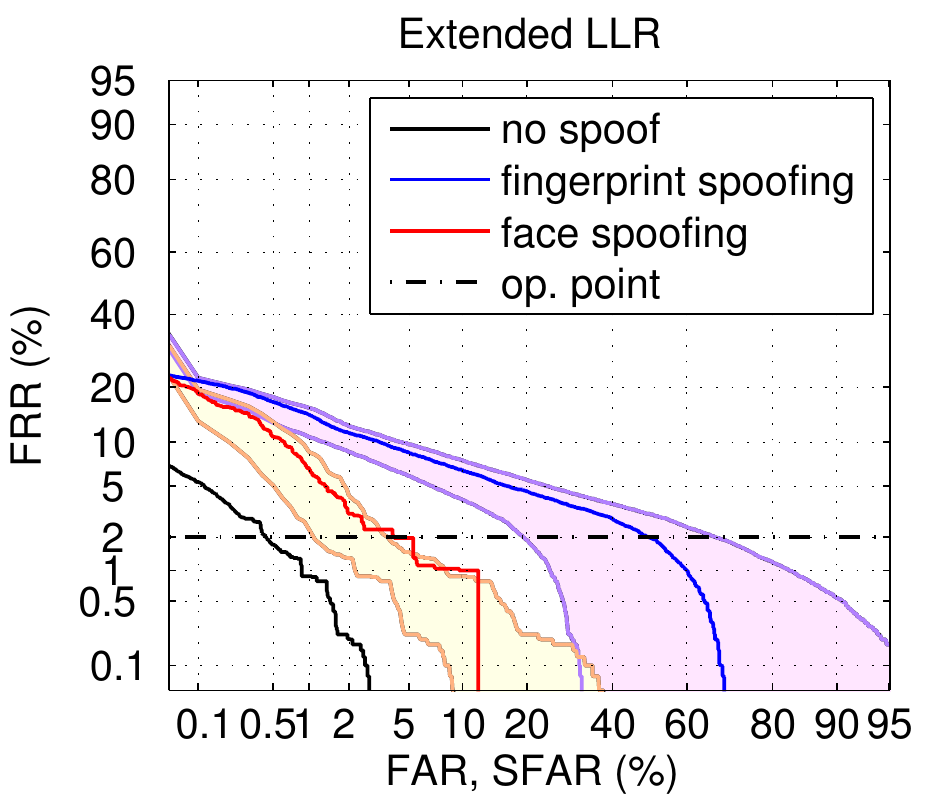}
\includegraphics[width = 0.22\textwidth]{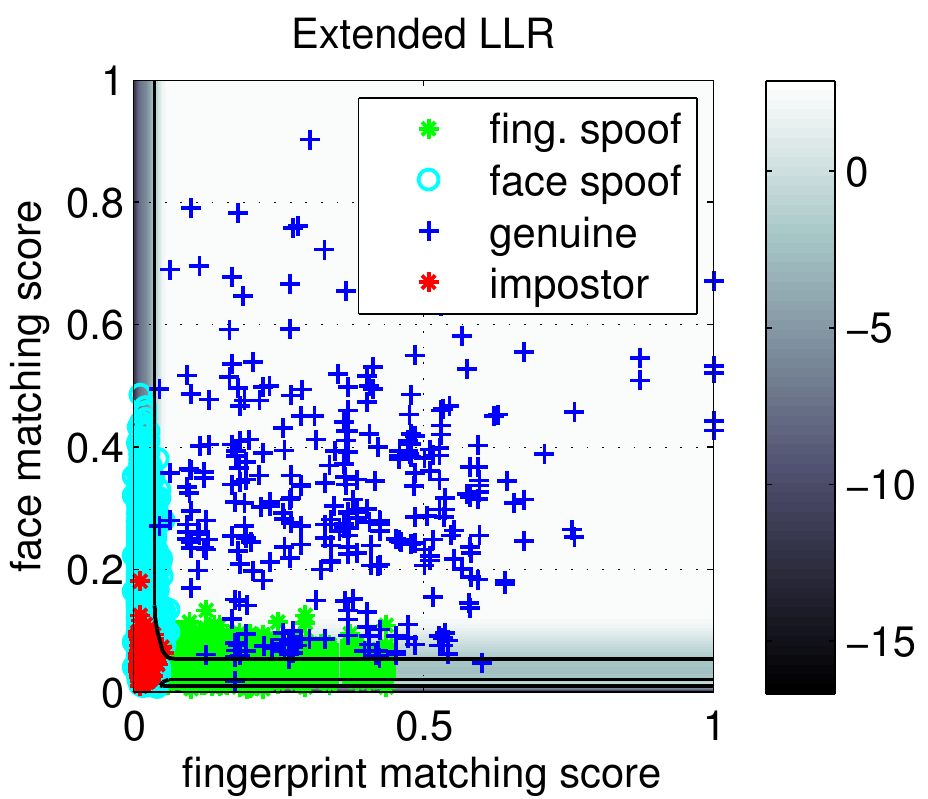}
\includegraphics[width = 0.23\textwidth]{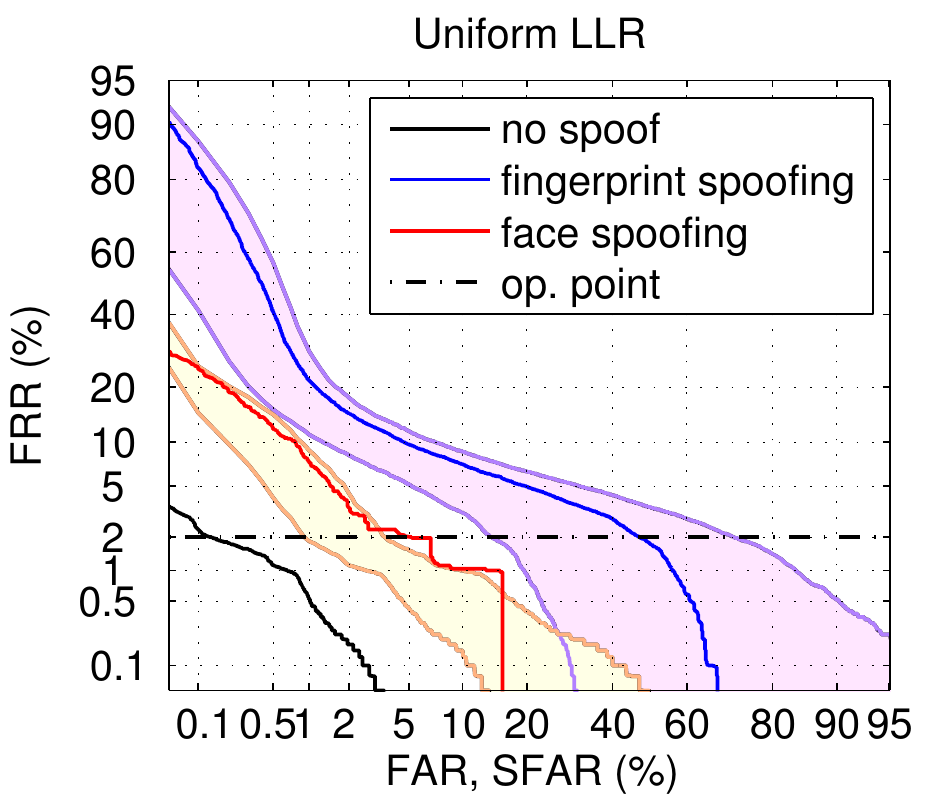}
\includegraphics[width = 0.22\textwidth]{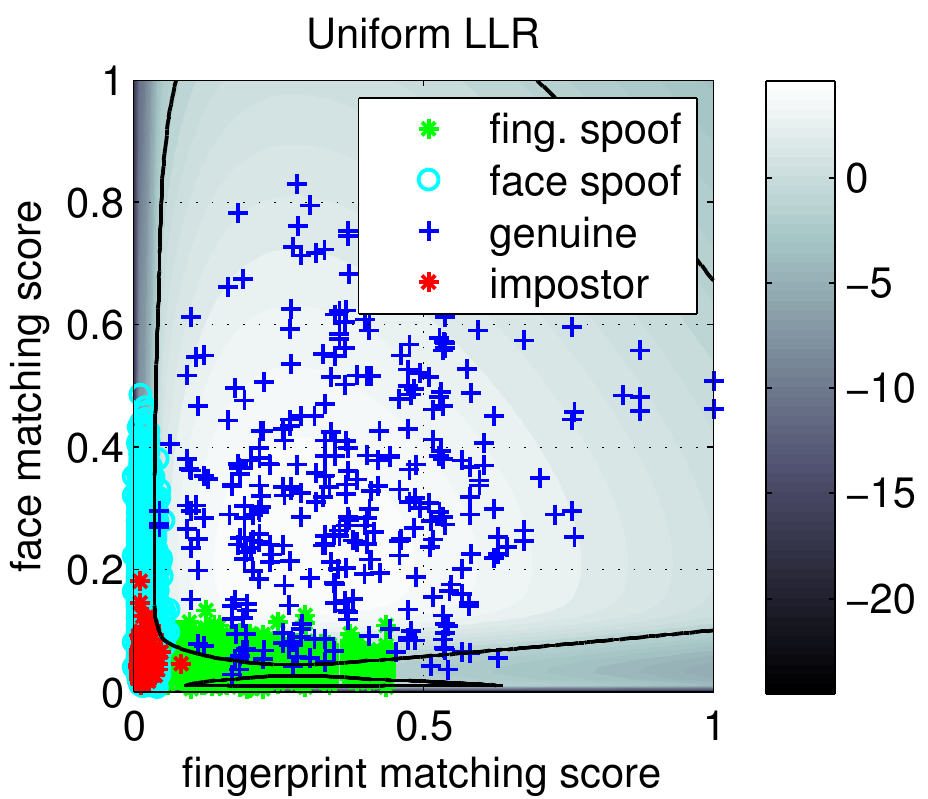}
\includegraphics[width = 0.23\textwidth]{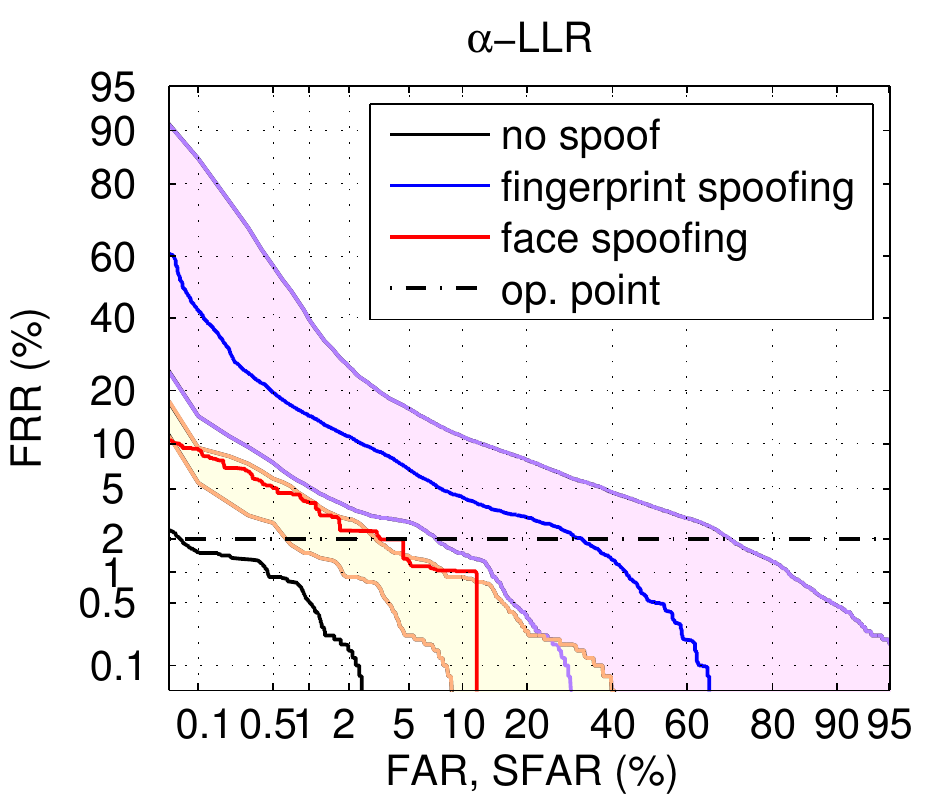}
\includegraphics[width = 0.22\textwidth]{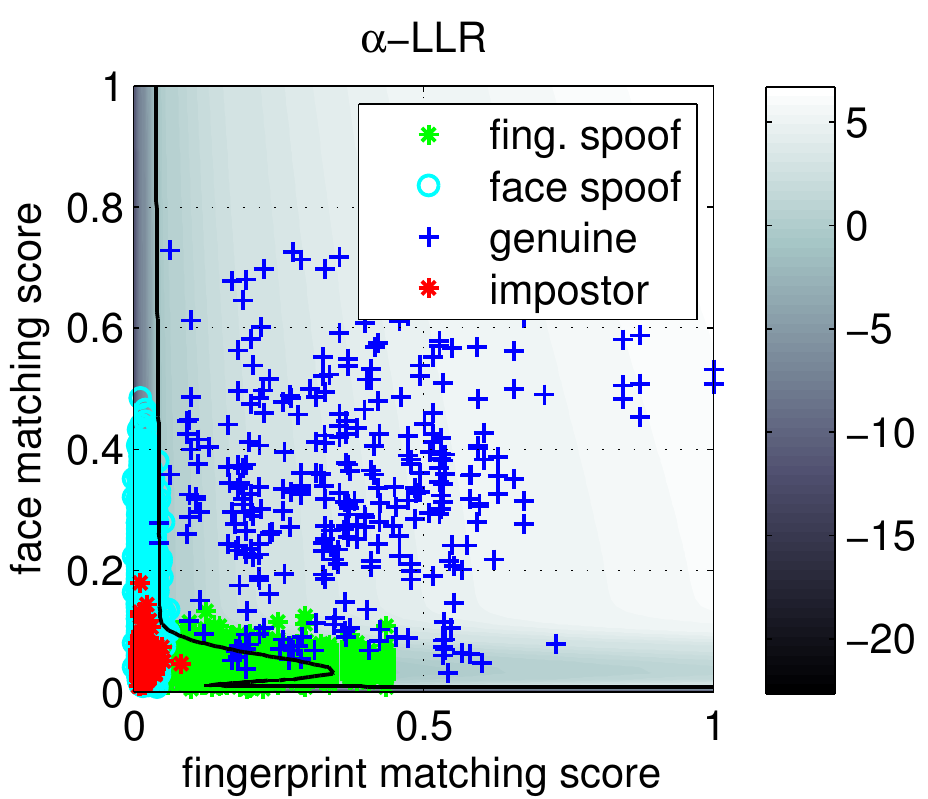}
\includegraphics[width = 0.23\textwidth]{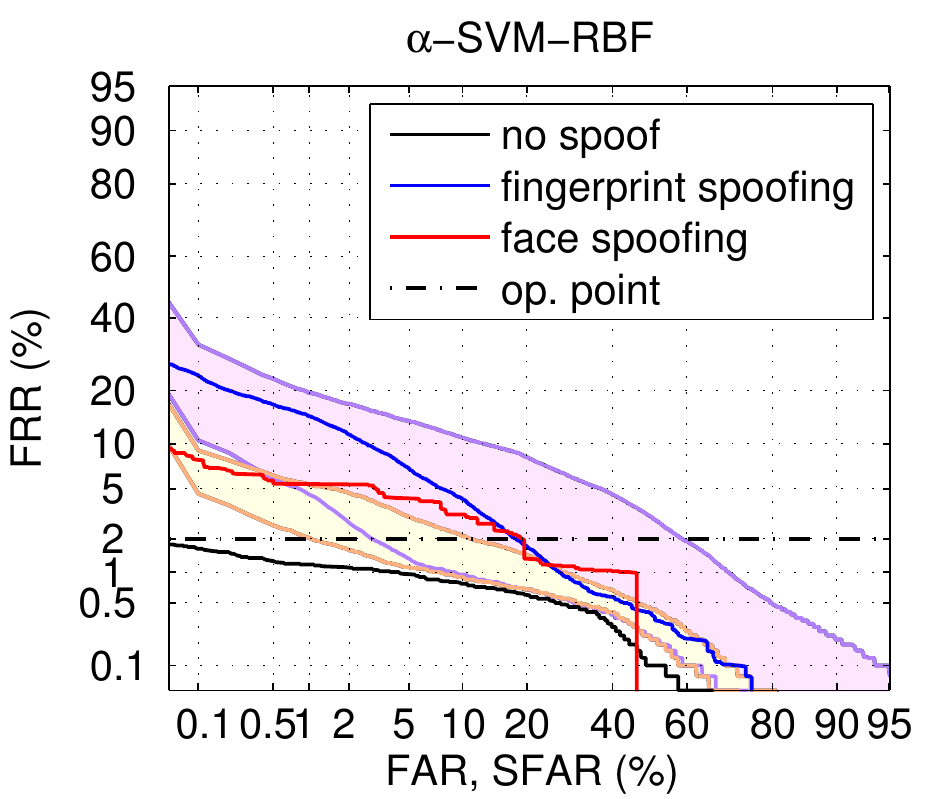}
\includegraphics[width = 0.22\textwidth]{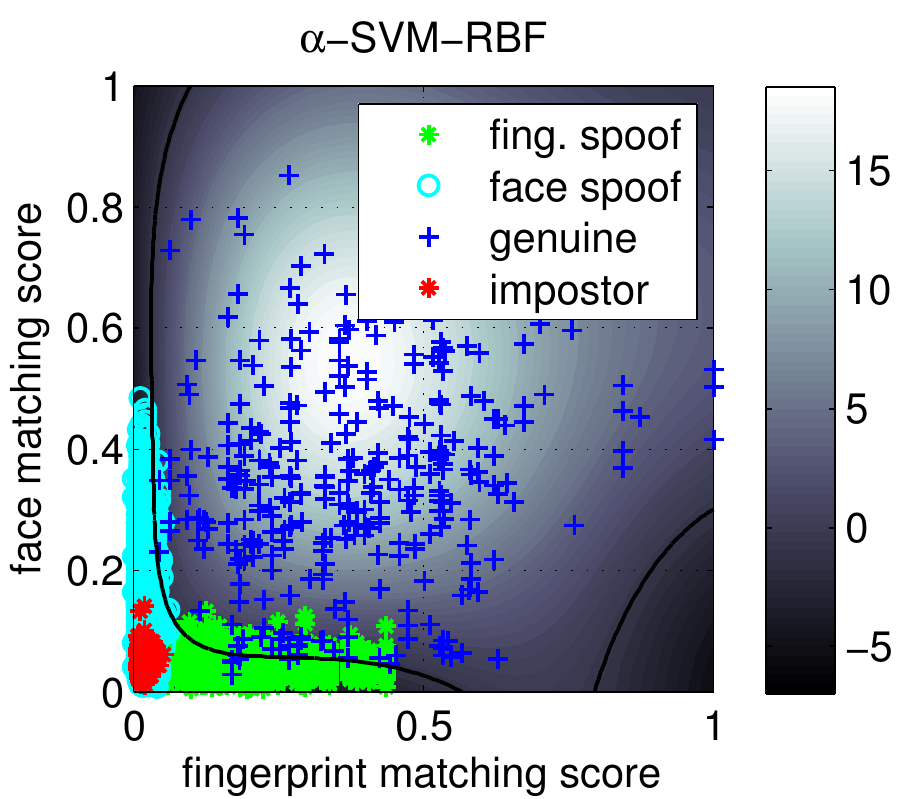}\\
\vspace{-8pt}
\caption{Results for the considered bimodal system. Plots in the first and third column report the average DET curves attained by each fusion rule, when no attack is considered (`no spoof'), and under presentation attacks from LivDet15 (`fingerprint spoofing') and CASIA  (`face spoofing'). The yellow and purple shaded areas represent the confidence bands for the ${\rm SFAR}$ predicted by our approach, over the family of fake score distributions represented by our meta-model, for face and fingerprint, respectively. The background color of plots in the second and fourth column represents the value of the fused score $f(\mathbf s)$ for each rule, in the space of matching scores. The black solid line represents its decision function at ${\rm FRR=2\%}$. We also report points corresponding to genuine users, impostors and presentation attacks, to compare the different decision functions.}
\label{fig:fing-face}
\end{center}
\vspace{-10pt}
\end{figure*}
\begin{table*}[ht]
\caption{Average (\%) performance (and standard deviation) attained by each rule at ${\rm FRR}=2\%$, in terms of ${\rm FAR}$, ${\rm SFAR}$ under the LivDet15 (${\rm SFAR}$ fing.) and CASIA (${\rm SFAR}$ face) presentation attacks, ${\rm SFAR}$ under the fingerprint (${\rm SFAR}_{\rm H1}$) and face (${\rm SFAR}_{\rm H2}$) \emph{high-impact} simulated attacks, and the corresponding ${\rm GFAR}$ values (denoted with ${\rm GFAR}$ for the LivDet15 and CASIA attacks, and ${\rm GFAR}_{\rm H1,H2}$ for the simulated attacks).}
\vspace{-10pt}
\centering
\resizebox{\textwidth}{!}{%
\begin{tabular}{|p{1.6cm}|p{1.295cm}|p{1.3cm}|p{1.295cm}|p{1.295cm}|p{1.295cm}|p{1.295cm}|p{1.295cm}|p{1.295cm}|p{1.295cm}|p{1.55cm}|}
\hline 
Rule & Sum & LDA & Product & Minimum & LLR & SVM-RBF & Ext. LLR & Unif. LLR & $\alpha$-LLR & $\alpha$-SVM-RBF \\ \hline 
${\rm FAR}$ & $0.0 \pm 0.0$ & $0.0 \pm 0.0$ & $0.0 \pm 0.0$ & $3.8 \pm 4.0$ & $0.0 \pm 0.0$ & $0.0 \pm 0.0$ & $0.4 \pm 0.4$ & $0.1 \pm 0.1$ & $0.1 \pm 0.1$ & $0.0 \pm 0.0$ \\ \hline
${\rm SFAR}$ fing. & $23.2 \pm 14.4$ & $25.8 \pm 16.2$ & $23.4 \pm 15.9$ & $53.5 \pm 25.6$ & $37.8 \pm 21.6$ & $42.0 \pm 20.6$ & $49.8 \pm 28.3$ & $47.4 \pm 27.2$ & $31.9 \pm 26.5$ & $17.9 \pm 12.2$ \\ 
${\rm SFAR}_{\rm H1}$ & $31.8 \pm 13.3$ & $33.7 \pm 13.4$ & $30.6 \pm 14.3$ & $51.6 \pm 11.8$ & $41.8 \pm 14.6$ & $46.1 \pm 14.2$ & $47.4 \pm 15.4$ & $45.9 \pm 15.0$ & $35.5 \pm 14.8$ & $25.0 \pm 7.6$ \\ \hline
${\rm SFAR}$ face & $15.3 \pm 14.0$ & $13.2 \pm 13.5$ & $4.9 \pm 6.9$ & $9.3 \pm 13.8$ & $3.1 \pm 4.6$ & $5.5 \pm 5.4$ & $4.0 \pm 7.4$ & $5.2 \pm 8.1$ & $3.3 \pm 7.4$ & $19.1 \pm 29.5$ \\ 
${\rm SFAR}_{\rm H2}$ & $39.5 \pm 11.6$ & $42.0 \pm 11.6$ & $37.9 \pm 11.9$ & $58.2 \pm 9.0$ & $51.8 \pm 12.2$ & $56.7 \pm 12.2$ & $55.1 \pm 12.4$ & $55.1 \pm 12.1$ & $43.0 \pm 12.0$ & $30.7 \pm 6.8$ \\  \hline
${\rm GFAR}$ & $9.6 \pm 6.1$ & $9.8 \pm 6.2$ & $\mathbf{7.1 \pm 4.8}$ & $17.6 \pm 9.3$ & $10.2 \pm 6.1$ & $11.9 \pm 6.2$ & $13.7 \pm 8.2$ & $13.2 \pm 8.1$ & $8.8 \pm 7.6$ & $9.3 \pm 9.4$ \\ 
${\rm GFAR_{\rm H1, H2}}$ & $17.8 \pm 4.4$ & $18.9 \pm 4.4$ & $17.1 \pm 4.7$ & $29.4 \pm 4.2$ & $23.4 \pm 4.8$ & $25.7 \pm 4.7$ & $25.8 \pm 5.0$ & $25.3 \pm 4.8$ & $19.7 \pm 4.8$ & $\mathbf{14.0 \pm 2.5}$ \\ 
\hline 
\end{tabular}
}
\vspace{-8pt}
\label{tab:FRR-fing-face}
\end{table*}

\textbf{SVM-RBF.} This rule consists of learning an SVM with the RBF kernel on the available training matching scores, to discriminate between genuine and impostor users.
We set the parameters $C \in \{ 0.001, 0.01, 0.1, 1, 10, 100 \}$ and  $\gamma \in \{ 0.01, 0.1, 1, 10, 100\}$ by minimizing the ${\rm FAR}$ at ${\rm FRR}=2\%$ through a 5-fold cross-validation on the training data. 

\textbf{Extended LLR.} This is the modified LLR proposed in \cite{rodrigues09}, as described in Sect.~\ref{sect:secure-fusion-old}. 
To minimize the ${\rm GFAR}$ according to Eqs.~\eqref{eq:exp-1}-\eqref{eq:exp-2}, we set the priors as $p(A_1=0, A_2=0 | {\rm I})=1/2$, $p(A_1=1, A_2=0 | {\rm I})=p(A_1=0, A_2=1 | {\rm I})=1/4$, and $p(A_1=1, A_2=1 | {\rm I})=0$, although this choice does not correspond to any specific choice of $r$, $c_{1}$ and $c_{2}$ for this rule. The fused matching score is given by Eq.~\eqref{eq:LLR}.

\textbf{Uniform LLR.}  This is the other LLR modification proposed in \cite{biggio11-smc}, and described in Sect.~\ref{sect:secure-fusion-old}. We set $p(A_1, A_2 | {\rm I})$ as for the Extended LLR, coherently with the given selection criterion. The combined score is given again by Eq.~\eqref{eq:LLR}

\textbf{$\alpha$-LLR.} For this rule too, we set the prior distribution $p(A_1, A_2 | {\rm I})$ as described for the Extended LLR and the Uniform LLR, in agreement with the selection criterion.
The distribution of attack samples $p(s_i | A_i = 1, Y={\rm I})$ is instead based on the \emph{high-risk} attack scenarios defined by our spoof simulation model. We therefore set $\mu_\alpha=0.40$ and $\sigma_\alpha=0.26$ for simulating attacks against the fingerprint matcher (\ie, when `RI' is used), and  $\mu_\alpha=0.91$ and $\sigma_\alpha=0.11$ for the face matcher (\ie, when `G' is used), according to Table~\ref{tab:model-parameters}. The fused score is given by Eq.~\eqref{eq:LLR}.

\textbf{$\alpha$-SVM-RBF.} We train this fusion rule using a modified training set sampled from the same distribution hypothesized for the $\alpha$-LLR, \ie, assuming the same prior $p(\vct A | {\rm I})$ and fake score distributions $p(s_i | A_i = 1, Y={\rm I})$. Such a training set can be obtained as explained in Sect.~\ref{sect:discriminative-approach}.
The values of $C$ and $\gamma$ are optimized using a 5-fold cross validation on the training data, as for the SVM-RBF, although here this amounts to minimizing the ${\rm GFAR}$ at ${\rm FRR}=2\%$, as the training data is modified according to the desired $p(\vct S | {\rm I})$. 

\begin{figure}[tb]
\begin{center}
\includegraphics[width = 0.24\textwidth]{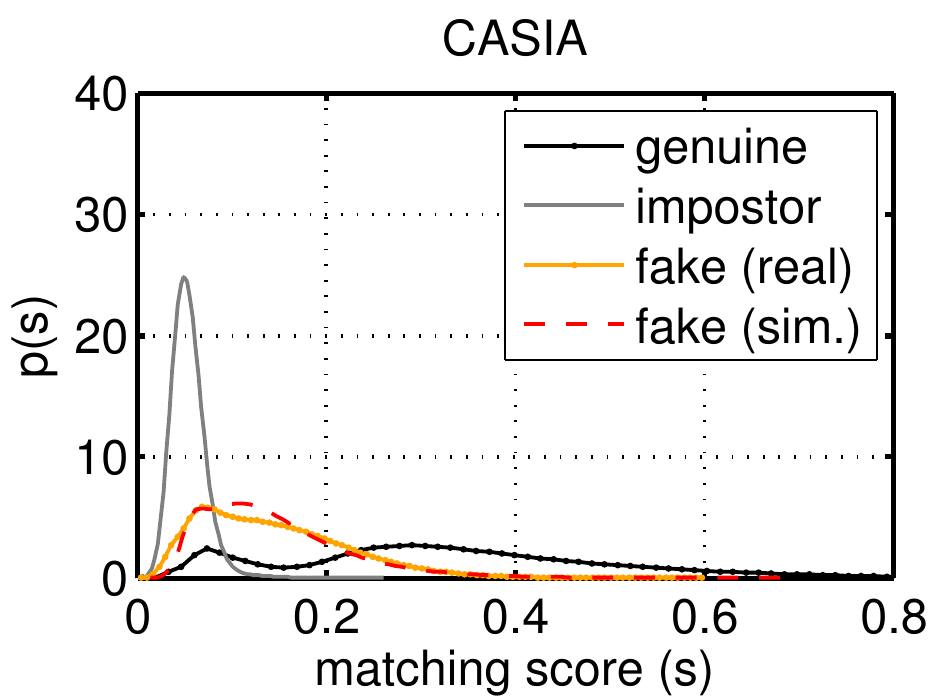}
\includegraphics[width = 0.24\textwidth]{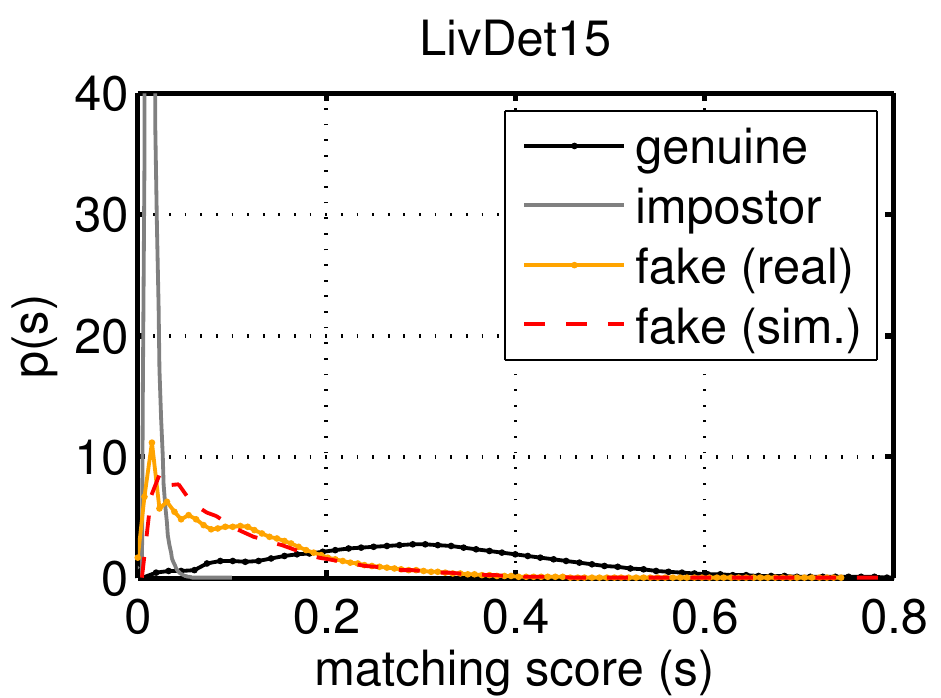}\\
\vspace{-10 pt}
\caption{Matching score distributions for CASIA and LivDet15. Fake score distributions fitted with our meta-model are shown for comparison. The values of $(\mu_{\alpha},\sigma_{\alpha})$ found to simulate them are $(0.33,0.15)$ for CASIA (\emph{attack impact}=$14\%$), and $(0.32,0.23)$ for LivDet15 (\emph{attack impact}=$23\%$).}
\label{fig:spoof-distributions-2}
\vspace{-15 pt}
\end{center}
\end{figure}

\subsection{Experimental Results} 
\label{sect:exp-results}

To show that our meta-model is capable of reliably modeling also the score distributions of never-before-seen presentation attacks, we first report in Fig.~\ref{fig:spoof-distributions-2} its fitting on the score distributions of the attacks included in LivDet15 and CASIA.
Note that the corresponding parameters $(\mu_{\alpha},\sigma_{\alpha})$
do not exactly match any of the attack scenarios defined in Table~\ref{tab:model-parameters} and Fig.~\ref{fig:model-fitting}. In fact, face and fingerprint spoofs from CASIA and LivDet15 exhibit an intermediate behavior respectively between the low- and med-impact attack scenarios for faces, and between the med- and high-impact attack scenarios for fingerprints. This further highlights that such attacks are very different from those considered in the design of our meta-model and of known attack scenarios.

The results for the given bimodal system are reported in terms of average Detection Error Trade-off (${\rm DET}$) curves in Fig.~\ref{fig:fing-face}. These curves report ${\rm FRR}$ vs ${\rm FAR}$ (or ${\rm SFAR}$) for all operating points on a convenient axis scaling~\cite{martin97,johnson10}.
Using our-meta model, we construct a family of ${\rm DET}$ curves, each obtained by simulating an attack scenario (\ie, a fake score distribution) against a single matcher.
We then average ${\rm DET}$ curves corresponding to attack scenarios that exhibit a similar \emph{attack impact}, yielding 20 distinct curves corresponding to the attack impact values $\{0,0.05,0.1,\ldots,1\}$.
The area covered by such ${\rm DET}$ curves, for each matcher, is highlighted using a shaded area, as described in Fig.~\ref{fig:fing-face}. 
We also report the ${\rm DET}$ curves corresponding to presentation attacks from the LivDet15 and CASIA databases.

To correctly understand our evaluation, recall that spoofing does not affect the matching score distribution of genuine users, \ie, the ${\rm FRR}$ under attack does not change.
Accordingly, for any operating point on the ${\rm DET}$ curve computed without attacks (corresponding to the `no spoof' scenario, which reports ${\rm FRR}$ vs ${\rm FAR}$), the ${\rm SFAR}$ values predicted by our model can be found by intersecting the shaded areas in Fig.~\ref{fig:fing-face} with the horizontal line corresponding to the same ${\rm FRR}$ value (see, \eg, `op. point' in Fig.~\ref{fig:fing-face}).

To provide a clearer discussion of our results, in Table~\ref{tab:FRR-fing-face}, we also report the average performance attained by each rule at the operating point corresponding to ${\rm FRR}=2\%$, including ${\rm FAR}$, and ${\rm SFAR}$ and ${\rm GFAR}$ (Eqs.~\ref{eq:exp-1}-\ref{eq:exp-2}) attained under the LivDet15 and CASIA attacks, and under the high-impact attacks simulated with our meta-model (see Table~\ref{tab:model-parameters}).

Let us first compare the predictions provided by our analysis against those corresponding to the  LivDet15 and CASIA presentation attacks.
As one may note from Fig.~\ref{fig:fing-face}, the confidence bands denoting the variability of the ${\rm DET}$ curves obtained under the attacks simulated with our meta-model almost always correctly represent and follow the behavior of the ${\rm DET}$ curves corresponding to the LivDet15 and CASIA attacks. This shows that our approach may be able to reliably predict the performance of a multibiometric system even under never-before-seen presentation attacks.

Furthermore, our analysis also highlights whether the fusion rule is more sensitive to variations in the output of a given matcher. This can be noted by comparing the confidence bands corresponding to  attacks against the fingerprint and the face matcher in Fig.~\ref{fig:fing-face}. In our case, fusion rules are generally more vulnerable to attacks targeting the fingerprint matcher (except for Sum and LDA), as the confidence bands corresponding to fingerprint presentation attacks are typically more shifted towards higher error rates. The reason is simply that the fingerprint matcher is more accurate than the face one in this case, and, thus, when the former is under attack, the matching scores of spoof impostors and genuine users tend to overlap more (\cf{} the scatter plots in the second and fourth column of Fig.~\ref{fig:fing-face}).

From the ${\rm DET}$ curves in Fig.~\ref{fig:fing-face}, one may also note that standard rules are generally more accurate in the absence of attack than secure fusion rules, confirming the trade-off between the performance in the absence and in the presence of spoofing.
The Minimum is an exception, as it exhibits a higher ${\rm FAR}$. 
The reason is that this rule only accepts a genuine claim if all the combined scores are sufficiently high. Thus, to keep an acceptable, low ${\rm FRR}$, one has to trade for a higher ${\rm FAR}$, and this may also worsen security against spoofing, conversely to intuition.

From Table~\ref{tab:FRR-fing-face}, one may also appreciate that secure fusion rule designed under the too pessimistic assumption given by Eq.~\eqref{eq:assumption-rodrigues} perform worse than the $\alpha$-LLR and the $\alpha$-SVM-RBF under the LivDet15 and CASIA presentation attacks.
This shows that our meta-model can also lead one to design fusion rules with an improved trade-off between the performance in the absence of attack and system security.

Besides giving a general overview of the kind of analysis enabled by our security evaluation procedure, the goal of the proposed case study is to show how a system designer can select a suitable fusion rule.
According to the criterion given by Eqs.~\eqref{eq:exp-1}-\eqref{eq:exp-2}, it is clear from Table~\ref{tab:FRR-fing-face} that the rule that attains the minimum expected ${\rm GFAR}$ (according to our model, the ${\rm GFAR}_{\rm H1,H2}$ value) is the $\alpha$-SVM-RBF, which can be thus selected as the fusion rule for this task. In particular, this rule attains also the lowest ${\rm SFAR}$ under the (simulated) fingerprint presentation attack (${\rm SFAR}_{\rm H1}$).
Notably, also Sum, LDA, Product and $\alpha$-LLR may be exploited to the same end, as they achieve only a slightly higher ${\rm GFAR}$. 

When considering the ${\rm GFAR}$ attained under the LivDet15 and CASIA spoofing attacks, the best rule turns out to be the Product rule, immediately followed by the $\alpha$-LLR and the $\alpha$-SVM-RBF. This is somehow reasonable to expect, as our analysis did not exploit any specific knowledge of such attacks, and, in particular, as we tuned our fusion rules using slightly overly-pessimistic attack settings (the fingerprint and face attack scenarios considered to design the $\alpha$-LLR and the $\alpha$-SVM-RBF have a higher \emph{attack impact} than that exhibited by the LivDet15 and CASIA attacks). Despite this, selecting the $\alpha$-SVM-RBF instead of the Product   would not raise any severe security issue in this case. 
Conversely, it may be even beneficial if fingerprint spoofing is deemed more likely during system operation than face spoofing. This should be clearly noted by the system designer before taking the final decision.

\textbf{Why Secure Fusion Works.} The fact that the standard fusion rules like the Product can be competitive with secure fusion rules in the presence of spoofing can be easily understood by looking at the shape of the decision functions in the scatter plots of Fig.~\ref{fig:fing-face}. 
In fact, the decision functions of such rules tend to better \emph{enclose} the genuine class rather than the impostor class.
Whereas untrained rules like the Product may perform well only under specific data distributions (like those shown in the depicted cases), trained secure fusion rules are expected to perform better on a wider variety of cases, due to their flexibility in learning and shaping the decision function depending on the given set of scores. However, providing a better enclosing of the genuine class turns out to clearly increase the ${\rm FAR}$ at the same ${\rm FRR}$ value (or vice versa), underlining again the trade-off between the performance in the absence of attacks and that under attack.
For this reason, it is especially important to be able to tune this trade-off properly, and not in an overly-pessimistic manner, as demonstrated in the design of the secure fusion rules based on our meta-model.

\section{Conclusions and Future Work}
\label{sect:conclusions}

We proposed an approach to thoroughly assess the security of multibiometric systems against presentation attacks, and to improve their security by design, overcoming the limitations of previous work~\cite{rodrigues09,rodrigues10,johnson10}.
Our approach is grounded on a statistical meta-model that incorporates knowledge of state-of-the-art fingerprint and face presentation attacks, by simulating their matching score distributions at the output of the attacked matchers, avoiding the cumbersome task of fabricating a large, representative set of attacks during system design.
It also allows us to simulate perturbations of such distributions that may correspond to \emph{unknown} attacks of different impact, through an \emph{uncertainty} analysis.
This aspect is specifically important, as attackers constantly aim to find novel 
evasion techniques~\cite{biggio14-tkde}.
In the case of biometric systems, this means that novel, unexpected attacks may be encountered in the near future. For instance, in~\cite{hadid14-cvprw}, it has been claimed that it is not possible to forecast all potential face spoofing attacks and fake fabrication techniques, as humans can always find very creative ways to cheat a system.
Our uncertainty analysis aims thus to overcome this issue. We showed empirically that our approach provides a much more informative security evaluation of multibiometric systems, characterizing the behavior of the system also under never-before-seen attacks, and enabling the design of improved secure fusion rules. 

We argue that our statistical meta-model can be applied to presentation attacks targeting other biometric traits, like palm vein and iris, as preliminary empirical evidences show that
their score distributions exhibit similar characteristics to those observed for face and fingerprint in Sect.~\ref{sect:alpha-model} (see, \eg, \cite{tome15-icb,busch15-tifs}). This however requires further investigation, and can be addressed in future work.

To conclude, it is also worth remarking that secure fusion may provide a complementary approach to liveness detection techniques that protect the combined matchers against spoofing. Accordingly, another interesting future extension of this work may be to exploit our meta-model in the context of recent approaches that combine liveness detection and matching algorithms, instead of using them as independent modules~\cite{marasco11-mcs,marasco12-btas,chingovska13,wild15-pr}. We believe that this may significantly improve multibiometric security to spoofing.




\begin{IEEEbiography}
[{\includegraphics[width=1in,height =1.25in,clip,keepaspectratio]{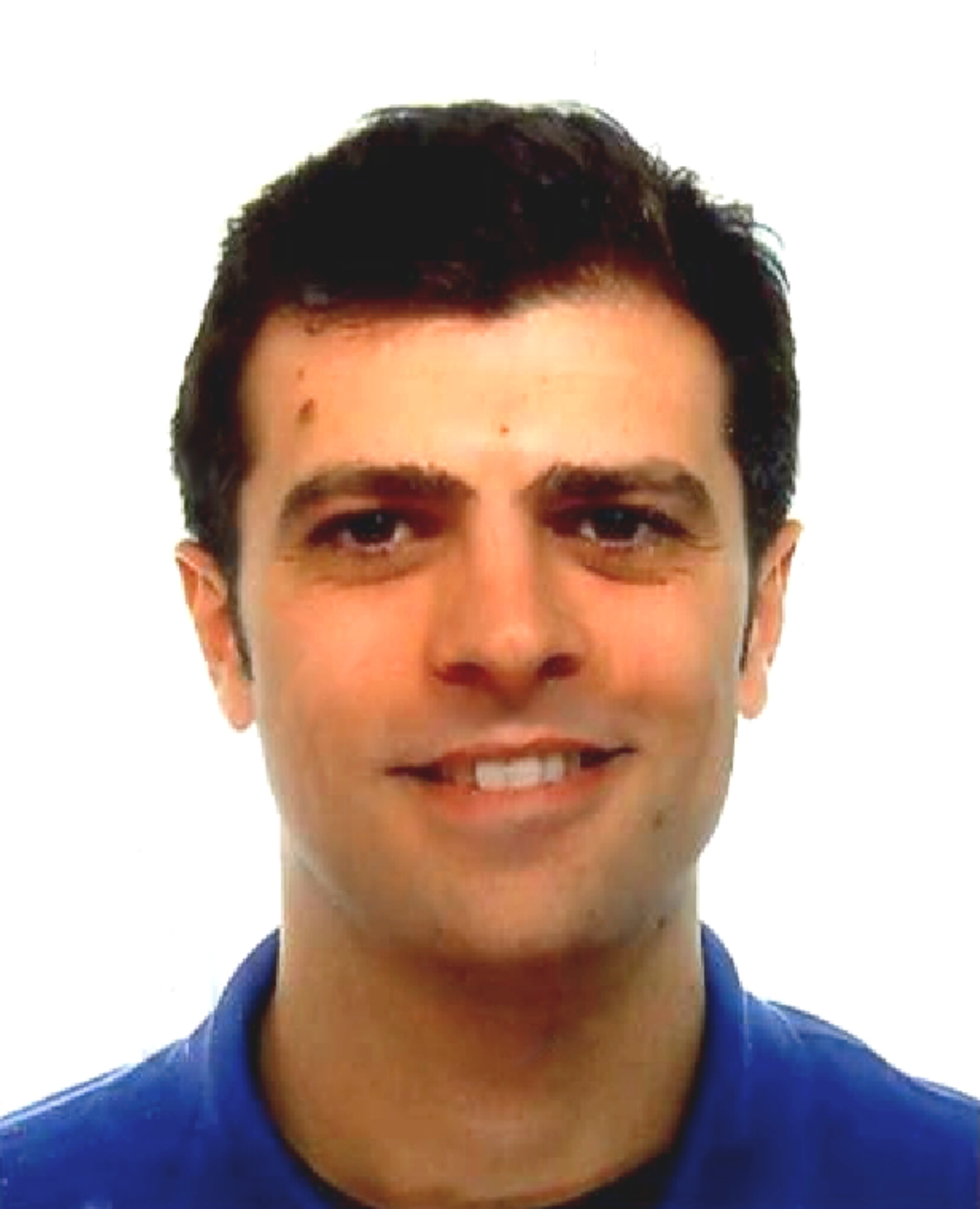}}]{Battista Biggio} received the M. Sc. degree in Electronic Eng., with honors, and the Ph. D. in Electronic Eng. and Computer Science, respectively in 2006 and 2010, from the University of Cagliari, Italy. Since 2007 he has been working for the Dept. of Electrical and Electronic Eng. of the same University, where he holds now a postdoctoral position. In 2011, he visited the University of T\"ubingen, Germany, and worked on the security of machine learning algorithms to contamination of training data. His research interests currently include: secure / robust machine learning and pattern recognition methods, multiple classifier systems, kernel methods, biometric authentication, spam filtering, and computer security. He serves as a reviewer for several international conferences and journals in the field. Dr. Biggio is a member of the IEEE (Computer Society, and Systems, Man and Cybernetics Society), and of the International Association for Pattern Recognition (IAPR).
\end{IEEEbiography}
\vspace{-5pt}

\begin{IEEEbiography}[{\includegraphics[width=1in,height =1.25in,clip,keepaspectratio]{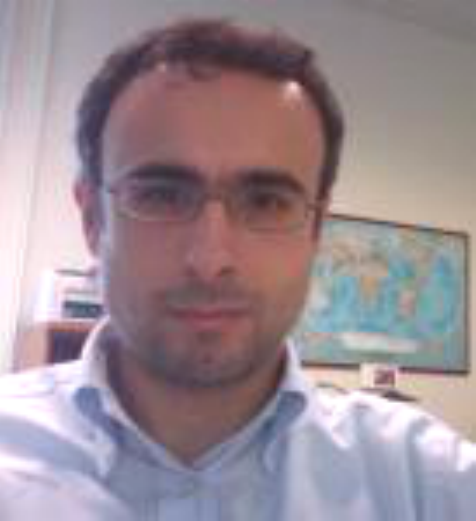}}]{Giorgio Fumera} received the M. Sc. degree in Electronic Eng., with honors, and the Ph.D. degree in Electronic Eng. and Computer Science, respectively in 1997 and 2002, from the University of Cagliari, Italy. Since February 2010 he is Associate Professor of Computer Eng. at the Dept. of Electrical and Electronic Eng. of the same University.
His research interests are related to methodologies and applications of statistical pattern recognition, and include multiple classifier systems, classification with the reject option, adversarial classification and document categorization. On these topics he published about eighty papers in international journal and conferences. He acts as reviewer for the main international journals in this field.
Dr. Fumera is a member of the IEEE (Computer Society), of the Italian Association for Artificial Intelligence (AI*IA), and of the IAPR.
\end{IEEEbiography}
\vspace{-10pt}

\begin{IEEEbiography}[{\includegraphics[width=1in,height =1.25in,clip,keepaspectratio]{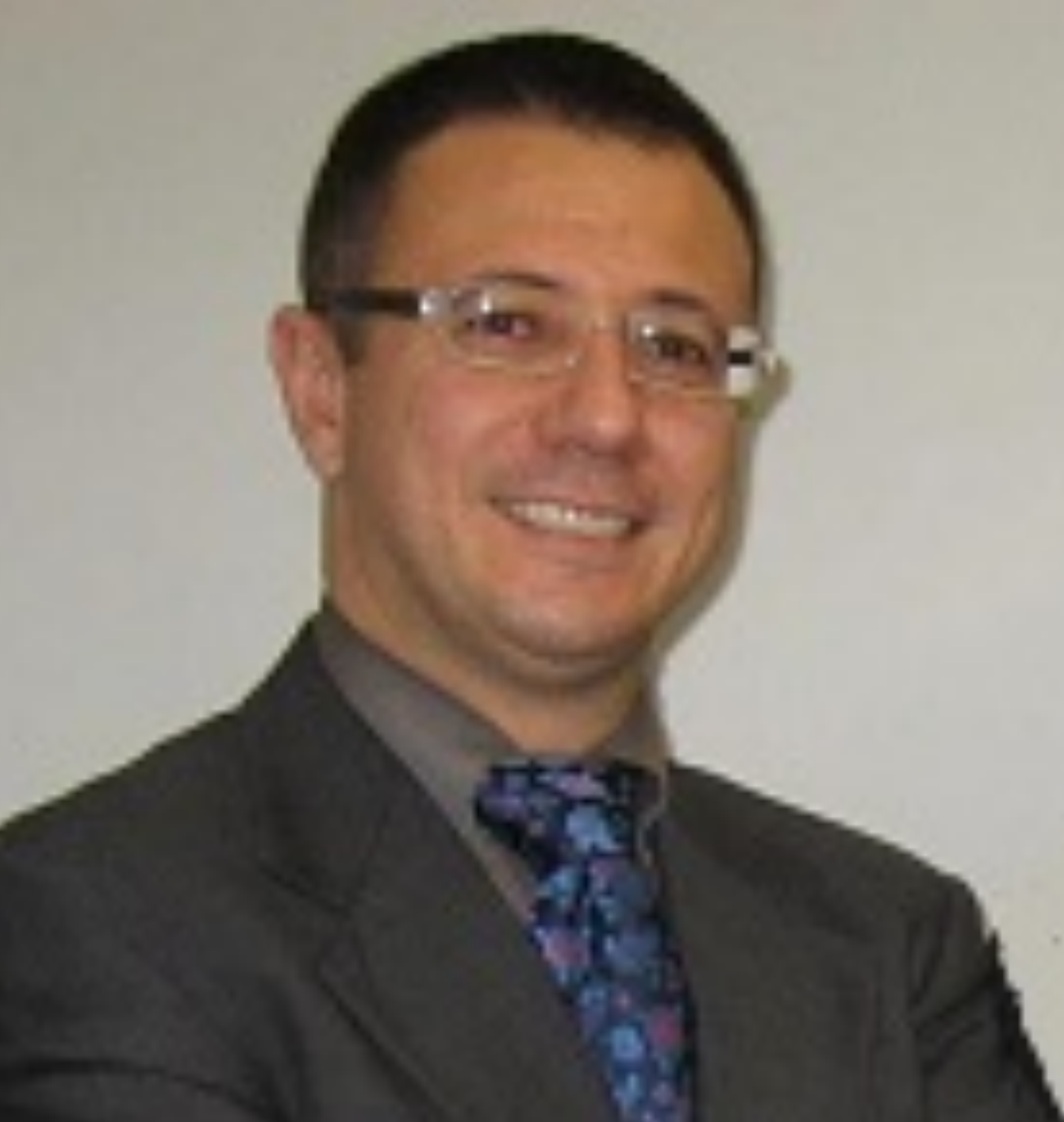}}]{Gian Luca Marcialis} received his M.Sc. degree and Ph.D. degree in Electronic and Computer Science Engineering from the University of Cagliari, Italy, in 2000 and 2004, respectively. He is currently Assistant Professor at the University of Cagliari. 
His research interests are in the field of biometrics, including fingerprint liveness detection, fingerprint classification and recognition, face recognition and head pose estimation, biometric template update by semi-supervised approaches, fusion of multiple biometric matchers for person recognition. On these topics, he co-authored more than 70 papers. He is co-organizer of the International Fingerprint Liveness Detection Competitions held in 2009, 2011, 2013 and 2015. Dr. Marcialis acts as referee for prestigious international journals and conferences, and is Editorial Review Board Member of International Journal of Digital Crime and Forensics. He is member of the IEEE and of the IAPR.
\end{IEEEbiography}
\vspace{-10pt}
\vfill
\begin{IEEEbiography}[{\includegraphics[width=1in,height =1.25in,clip,keepaspectratio]{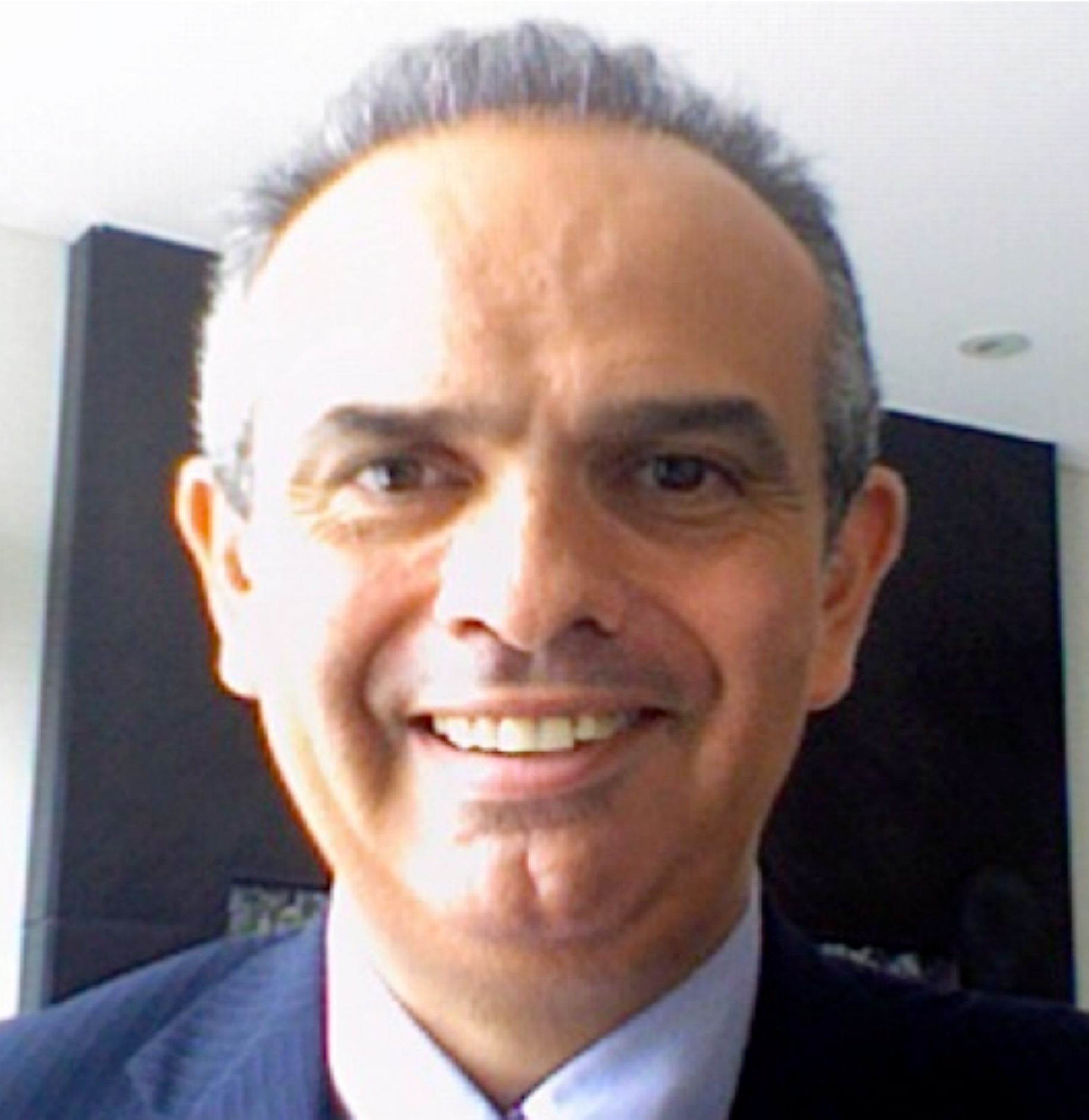}}]{Fabio Roli} received his M. Sc. degree, with honors, and Ph. D. degree in Electronic Eng. from
the University of Genoa, Italy.
He is professor of computer engineering
and head of the research laboratory on pattern
recognition and applications at the Dept. of Electrical and
Electronic Eng. of the University of Cagliari, Italy,
His research activity is focused on the
design of pattern recognition systems and their applications to biometric
personal identification, multimedia text categorization, and computer
security. On these topics, he has published more than two hundred
papers at conferences and on journals. He was a very active organizer of
international conferences and workshops, and established the popular
workshop series on multiple classifier systems. Dr. Roli is a member
of the governing boards of the IAPR and of the IEEE Systems, Man and Cybernetics Society.
He is Fellow of the IEEE, and Fellow of the IAPR.
\end{IEEEbiography}

\end{document}